\documentclass[9pt,twoside,openany]{book}
\usepackage[a4paper, margin=0.9in, top=1in, bottom=1in]{geometry}
\usepackage{tikz}
\usepackage{amsmath}
\usepackage{amssymb}
\usepackage{amsthm}
\usepackage{algorithm}
\usepackage{algpseudocode}
\usepackage{xcolor}
\usepackage{graphicx}
\usepackage{booktabs}
\usepackage{caption}
\usepackage{subcaption}
\usepackage{hyperref}
\usepackage{fancyhdr}
\usepackage{lipsum}
\usepackage{enumitem}
\usepackage{tcolorbox}
\usepackage{titlesec}
\usepackage{titletoc}
\usepackage{multicol}

\usetikzlibrary{arrows.meta, positioning, shapes, calc, decorations.pathreplacing, backgrounds, fit}

\definecolor{anthropicOrange}{RGB}{230, 120, 60}
\definecolor{anthropicBeige}{RGB}{245, 235, 220}
\definecolor{anthropicBrown}{RGB}{140, 100, 70}
\definecolor{anthropicCream}{RGB}{255, 250, 240}
\definecolor{anthropicTan}{RGB}{210, 180, 140}
\definecolor{anthropicDeepOrange}{RGB}{200, 90, 40}
\definecolor{anthropicLightOrange}{RGB}{255, 200, 150}

\usepackage{amsthm}

\theoremstyle{definition}
\newtheorem{theorem}{Theorem}[chapter]

\newtheorem{proposition}[theorem]{Proposition}
\newtheorem{corollary}[theorem]{Corollary}  
\newtheorem{definition}{Definition}[chapter]
\newtheorem{example}{Example}[chapter]

\newtcolorbox{keypoint}{
	colback=anthropicBeige,
	colframe=anthropicOrange,
	fonttitle=\bfseries,
	title=Key Points,
	arc=3mm,
	boxrule=1.5pt
}

\newtcolorbox{insight}{
	colback=anthropicCream,
	colframe=anthropicBrown,
	fonttitle=\bfseries,
	title=Mathematical Insight,
	arc=3mm,
	boxrule=1.5pt
}

\titleformat{\chapter}[display]
{\normalfont\Large\bfseries\color{anthropicDeepOrange}}
{\chaptertitlename\ \thechapter}{20pt}{\LARGE}
\titlespacing*{\chapter}{0pt}{0pt}{40pt}

\titleformat{\section}
{\normalfont\normalsize\bfseries\color{anthropicOrange}}
{\thesection}{1em}{}

\titleformat{\subsection}
{\normalfont\small\bfseries\color{anthropicOrange}}
{\thesubsection}{1em}{}

\titlecontents{chapter}
[0em]
{\addvspace{1em}\small\bfseries\color{anthropicBrown}}
{\color{anthropicBrown}\contentslabel{2em}}
{\color{anthropicBrown}\hspace*{-2em}}
{\titlerule*[0.5pc]{.}\contentspage}

\renewcommand{\headrulewidth}{0.5pt}
\renewcommand{\headrule}{\hbox to\headwidth{\color{anthropicOrange}\leaders\hrule height \headrulewidth\hfill}}

\hypersetup{
	colorlinks=true,
	linkcolor=anthropicOrange,
	citecolor=anthropicBrown,
	urlcolor=anthropicDeepOrange
}

\title{\Huge\textbf{Attention Mechanisms in Neural Networks}\\[0.5em]
	\Large A Comprehensive Mathematical Treatment\\[1em]
	\large From Theory to Implementation}
\author{\Large\textbf{Hasi Hays}}
\date{\today}

\begin{document}
	\frontmatter
	
\begin{titlepage}
	\begin{tikzpicture}[remember picture, overlay]
		\fill[anthropicBeige] (current page.south west) rectangle (current page.north east);

		\begin{scope}[opacity=0.15]
			\foreach \i in {1,...,5} {
				\fill[anthropicBrown] ($(current page.west) + (3cm, 5cm + \i*1.5cm)$) circle (0.3cm);
			}

			\foreach \i in {1,...,5} {
				\fill[anthropicOrange] ($(current page.center) + (-4cm, 5cm + \i*1.5cm)$) circle (0.25cm);
				\fill[anthropicOrange] ($(current page.center) + (-1cm, 5cm + \i*1.5cm)$) circle (0.25cm);
				\fill[anthropicOrange] ($(current page.center) + (2cm, 5cm + \i*1.5cm)$) circle (0.25cm);
			}

			\foreach \i in {1,...,5} {
				\foreach \j in {1,...,5} {
					\draw[anthropicDeepOrange, line width=0.3mm, opacity=0.3]
					($(current page.center) + (-4cm, 5cm + \i*1.5cm)$) --
					($(current page.center) + (-1cm, 5cm + \j*1.5cm)$);
				}
			}

			\foreach \i in {1,...,5} {
				\fill[anthropicBrown] ($(current page.east) + (-3cm, 5cm + \i*1.5cm)$) circle (0.3cm);
			}

			\foreach \x in {0,...,4} {
				\foreach \y in {0,...,4} {
					\pgfmathsetmacro{\opacity}{0.1 + 0.15*rnd}
					\fill[anthropicOrange, opacity=\opacity]
					($(current page.north east) + (-6cm - \x*0.8cm, -5cm - \y*0.8cm)$)
					rectangle ++(0.7cm, 0.7cm);
				}
			}

			\node[anthropicBrown, scale=3, rotate=-15] at ($(current page.north) + (0, -4cm)$) {$\mathbf{Q}$};
			\node[anthropicBrown, scale=3, rotate=15] at ($(current page.north) + (3cm, -6cm)$) {$\mathbf{K}$};
			\node[anthropicBrown, scale=3, rotate=-10] at ($(current page.north) + (-3cm, -6cm)$) {$\mathbf{V}$};

			\node[anthropicOrange, scale=2, rotate=20] at ($(current page.south) + (-5cm, 21cm)$) {$\text{softmax}$};

			\node[anthropicBrown, scale=1.5, rotate=-5] at ($(current page.south) + (5cm, 20cm)$) {$\alpha_{ij}$};

		\end{scope}

		\node[align=center] at (current page.center) {
			{\fontsize{40}{48}\selectfont\textbf{\textcolor{anthropicDeepOrange}{Attention Mechanisms}}}\\[0.5cm]
			{\fontsize{35}{42}\selectfont\textbf{\textcolor{anthropicOrange}{in Neural Networks}}}\\[1.5cm]
			{\fontsize{14}{14}\selectfont\textcolor{anthropicBrown}{A Comprehensive Mathematical Treatment}}\\[0.5cm]
			{\fontsize{14}{14}\selectfont\textcolor{anthropicBrown}{From Theory to Implementation}}
		};

		\node[align=center] at ($(current page.south) + (0, 4cm)$) {
			{\fontsize{20}{30}\selectfont\textbf{\textcolor{anthropicDeepOrange}{Hasi Hays}}}
		};

	\end{tikzpicture}
\end{titlepage}
	
	\chapter*{Abstract}
	\addcontentsline{toc}{chapter}{Abstract}
	
	Attention mechanisms represent a fundamental paradigm shift in neural network architectures, enabling models to selectively focus on relevant portions of input sequences through learned weighting functions. This monograph provides a comprehensive and rigorous mathematical treatment of attention mechanisms, encompassing their theoretical foundations, computational properties, and practical implementations in contemporary deep learning systems. We begin by establishing the mathematical framework for attention operations, defining the core components of queries, keys, and values, and analyzing various scoring functions including additive, multiplicative, and scaled dot-product attention. The permutation equivariance property of self-attention is proven, and its implications for positional encoding are thoroughly examined. The Transformer architecture is presented with complete mathematical derivations of all components, including multi-head attention, positional encoding schemes, feed-forward networks, and layer normalization. We provide a detailed complexity analysis that demonstrates the computational cost and memory requirements $O(n^2d)$, explaining both the advantages and limitations of the standard attention mechanism. Training dynamics are explored through the lens of optimization theory, including learning rate scheduling, gradient flow analysis through residual connections, and regularization techniques specific to attention-based models. We derive the backpropagation equations for attention layers and analyze the gradient landscape properties that facilitate training of deep Transformer networks. Extensive coverage of attention variants addresses the quadratic complexity limitation through sparse attention patterns, linear approximations via kernel methods, and efficient architectures for long-sequence modeling. Each variant is analyzed for its computational complexity, memory requirements, and theoretical approximation properties.
	
	Applications in natural language processing, computer vision, and multimodal learning demonstrate the versatility of attention mechanisms. We examine language modeling with autoregressive transformers, bidirectional encoders for representation learning, sequence-to-sequence translation, Vision Transformers for image classification, and cross-modal attention for vision-language tasks. Empirical analysis reveals training characteristics, scaling laws that relate performance to model size and computation, attention pattern visualizations, and performance benchmarks across standard datasets. We discuss the interpretability of learned attention patterns and their relationship to linguistic and visual structures. The monograph concludes with a critical examination of current limitations, including computational scalability, data efficiency, systematic generalization, and interpretability challenges. Future research directions are identified in efficient architectures, theoretical understanding, continual learning, and multimodal integration. A comprehensive appendix provides detailed notation tables and mathematical conventions for quick reference. This work establishes attention mechanisms within the broader context of sequence modeling and representation learning, providing researchers and practitioners with a unified mathematical framework for understanding, implementing, and extending these fundamental components of modern artificial intelligence systems.
	
	\mainmatter

	\hypersetup{linkcolor=anthropicDeepOrange}
	\tableofcontents
	\hypersetup{linkcolor=anthropicOrange}
		
	\chapter{Introduction}
	
	\section{The Attention Paradigm}
	
	The capacity to selectively process information represents a fundamental aspect of intelligent systems, both biological and artificial. In human cognition, attention mechanisms allow us to focus on relevant stimuli while filtering distractions, enabling efficient information processing in complex environments. This principle has proven equally transformative in artificial neural networks. Traditional neural architectures for sequence processing, including recurrent neural networks (RNNs) and long short-term memory (LSTM) networks, process inputs sequentially while maintaining a fixed-dimensional hidden state vector. This state vector must compress all relevant historical information into a finite representation, creating an information bottleneck that limits the model's capacity to capture long-range dependencies and complex relationships across distant sequence positions.
	
	\begin{keypoint}
		Attention mechanisms fundamentally change how neural networks process sequential data by allowing direct access to all positions in a sequence, eliminating the need for compression into fixed-dimensional representations.
	\end{keypoint}
	
	The attention operation computes a weighted sum over input representations, where the weights reflect the relevance or importance of each input position to the current computation. Rather than forcing all information through a narrow bottleneck, attention creates multiple pathways for information flow, with the network learning which pathways to emphasize for each specific computation.
	
	\subsection{Information Flow in Neural Networks}
	
	Consider a sequence $\mathbf{x}_1, \mathbf{x}_2, \ldots, \mathbf{x}_n$ that must be processed to produce outputs $\mathbf{y}_1, \mathbf{y}_2, \ldots, \mathbf{y}_n$. In recurrent architectures, information flows sequentially:
	
	\begin{equation}
		\mathbf{h}_t = f(\mathbf{h}_{t-1}, \mathbf{x}_t)
	\end{equation}
	
	where $\mathbf{h}_t$ is the hidden state at position $t$ and $f$ is a learned function. The hidden state $\mathbf{h}_t$ must encode all relevant information from positions $1$ through $t$ that might be needed for future computations. This creates several challenges:
	
	\begin{enumerate}[leftmargin=*]
		\item \textbf{Vanishing Gradients}: During backpropagation through time, gradients must flow backwards through the sequential dependencies, potentially diminishing exponentially with sequence length.
		
		\item \textbf{Information Bottleneck}: The fixed dimensionality of $\mathbf{h}_t$ limits the amount of information that can be preserved from distant positions.
		
		\item \textbf{Sequential Processing}: Computing $\mathbf{h}_t$ requires $\mathbf{h}_{t-1}$, preventing parallel computation across time steps.
		
		\item \textbf{Long-Range Dependencies}: Relationships between distant positions must be maintained through intermediate hidden states, making them difficult to learn.
	\end{enumerate}
	
	Attention mechanisms address these limitations through a fundamentally different approach to information flow. Instead of sequential processing, attention allows each position to directly access information from all other positions through learned attention weights.
	
	\subsection{The Attention Operation}
	
	At its core, the attention mechanism computes a contextualized representation for each position by taking a weighted average of representations from all positions. The weights are determined by learned compatibility functions that measure the relevance between positions.
	
	For position $i$ attending to all positions $j = 1, \ldots, n$, the attention operation computes:
	
	\begin{align}
		e_{ij} &= \text{score}(\mathbf{q}_i, \mathbf{k}_j) \\
		\alpha_{ij} &= \frac{\exp(e_{ij})}{\sum_{j=1}^{n} \exp(e_{ij})} \\
		\mathbf{c}_i &= \sum_{j=1}^{n} \alpha_{ij} \mathbf{v}_j
	\end{align}
	
	where $\mathbf{q}_i$ is a query vector representing what information position $i$ seeks, $\mathbf{k}_j$ is a key vector representing what information position $j$ offers, and $\mathbf{v}_j$ is a value vector containing the actual information to be aggregated. The attention weights $\alpha_{ij}$ form a probability distribution over positions, determining how much each position contributes to the output at position $i$.
	
	This formulation provides several critical advantages:
	
	\begin{enumerate}[leftmargin=*]
		\item \textbf{Direct Information Pathways}: Position $i$ can directly access information from position $j$ regardless of distance, with attention weight $\alpha_{ij}$ controlling the information flow.
		
		\item \textbf{Parallel Computation}: All positions can compute their attention-weighted representations simultaneously, enabling efficient parallel processing.
		
		\item \textbf{Adaptive Routing}: The network learns which positions should attend to which other positions for each specific task, rather than using fixed connectivity patterns.
		
		\item \textbf{Interpretability}: The attention weights $\alpha_{ij}$ provide some interpretability, showing which positions the model considers relevant for each computation.
	\end{enumerate}
	
	\section{Historical Development and Key Innovations}
	
	\subsection{Early Attention Mechanisms}
	
	The concept of attention in neural networks emerged from research in neural machine translation. Sutskever et al.~\cite{sutskever2014sequence} demonstrated that sequence-to-sequence models using LSTM encoders and decoders could perform machine translation by encoding the source sentence into a fixed-dimensional vector and then decoding this vector into the target language. However, this approach struggled with long sentences due to the information bottleneck in the fixed-dimensional encoding. Bahdanau et al.~\cite{bahdanau2014neural} introduced the first attention mechanism to address this limitation. Their innovation allowed the decoder to selectively focus on different parts of the source sentence when generating each target word.
	
	The Bahdanau attention mechanism computes alignment scores between the decoder's current state $\mathbf{s}_t$ and each encoder hidden state $\mathbf{h}_j$:
	
	\begin{equation}
		e_{tj} = \mathbf{v}_a^T \tanh(\mathbf{W}_a\mathbf{s}_t + \mathbf{U}_a\mathbf{h}_j)
	\end{equation}
	
	These scores are normalized via softmax to produce attention weights, which are used to compute a context vector as a weighted sum of encoder states. This context vector augments the decoder's input, providing targeted information from the source sequence.
	
	\subsection{Multiplicative Attention}
	
	Luong et al.~\cite{luong2015effective} proposed simplified attention mechanisms that replaced the additive scoring function with multiplicative variants:
	
	\begin{equation}
		e_{tj} = \mathbf{s}_t^T \mathbf{W} \mathbf{h}_j
	\end{equation}
	
	or, when dimensions align, simple dot-product attention:
	
	\begin{equation}
		e_{tj} = \mathbf{s}_t^T \mathbf{h}_j
	\end{equation}
	
	These multiplicative attention mechanisms proved computationally more efficient while maintaining effectiveness, particularly when combined with proper initialization and normalization schemes.
	
	\subsection{Self-Attention and the Transformer}
	
	The most significant breakthrough came with the Transformer architecture~\cite{vaswani2017attention}, which made two key innovations:
	
	\begin{enumerate}[leftmargin=*]
		\item \textbf{Self-Attention as Primary Operation}: Rather than using attention to augment recurrent or convolutional networks, the Transformer constructs models entirely from attention mechanisms.
		
		\item \textbf{Multi-Head Attention}: Multiple attention functions operate in parallel, each potentially learning to capture different types of relationships between positions.
	\end{enumerate}
	
	The Transformer's self-attention mechanism allows each position in a sequence to attend to all positions in the same sequence, including itself. This enables the model to capture complex dependencies without the sequential bottleneck of recurrent architectures.
	
	The scaled dot-product attention used in Transformers addresses numerical stability issues:
	
	\begin{equation}
		\text{Attention}(\mathbf{Q}, \mathbf{K}, \mathbf{V}) = \text{softmax}\left(\frac{\mathbf{Q}\mathbf{K}^T}{\sqrt{d_k}}\right)\mathbf{V}
	\end{equation}
	
	The scaling factor $1/\sqrt{d_k}$ prevents the dot products from growing large in magnitude as dimensionality increases, which would otherwise push the softmax into regions with vanishingly small gradients.
	
	\subsection{Impact and Adoption}
	
	The Transformer architecture has fundamentally transformed multiple domains:
	
	\textbf{Natural Language Processing}: Models like BERT~\cite{devlin2019bert}, GPT~\cite{radford2019language}, and T5~\cite{raffel2020exploring} have achieved state-of-the-art performance across virtually all NLP benchmarks, from question answering to text generation.
	
	\textbf{Computer Vision}: Vision Transformers~\cite{dosovitskiy2020image} demonstrated that attention mechanisms can match or exceed convolutional neural networks for image classification when trained on sufficient data, challenging the dominance of CNNs in computer vision.

	\textbf{Multimodal Learning}: Attention enables effective fusion of information across modalities, with models like CLIP~\cite{radford2021learning} learning joint representations of images and text through contrastive training.
	
	\textbf{Scientific Applications}: Attention-based models have been applied to protein folding with AlphaFold~\cite{jumper2021highly}, drug discovery, genomics, molecular biology~\cite{hays2025hierarchical}, and other scientific domains requiring complex pattern recognition and relationship modeling.
	
	\section{Mathematical Foundations Preview}
	
	The subsequent chapters develop the rigorous mathematical framework underlying attention mechanisms. We establish notation and definitions, prove key properties, analyze computational complexity, and derive training algorithms.
	
	Chapter 2 presents the mathematical foundations, defining the attention function formally and establishing its properties. We prove that self-attention is permutation equivariant, analyze different scoring functions, and derive the vectorized formulation used in efficient implementations.
	
	Chapter 3 focuses on self-attention mechanisms specifically, proving complexity results and analyzing the learned attention patterns. We demonstrate that self-attention has $O(n^2d)$ complexity and discuss implications for long sequence processing.
	
	Chapter 4 extends the analysis to multi-head attention, proving that multiple heads can learn complementary patterns and analyzing the representation capacity of multi-head architectures.
	
	Chapter 5 presents the complete Transformer architecture, deriving all components, including positional encodings, feed-forward networks, and layer normalization. We prove properties of sinusoidal positional encodings and analyze the role of residual connections.
	
	Subsequent chapters cover training dynamics, computational complexity, architectural variants, applications, empirical results, and future directions.
	
	\section{Scope and Organization}
	
	This monograph aims to provide a complete mathematical treatment of attention mechanisms, from first principles through state-of-the-art applications. We assume familiarity with linear algebra, calculus, probability theory, and basic neural network concepts, but derive all attention-specific results in detail. A comprehensive table of notation and mathematical conventions is provided in \autoref{app:notation} for quick reference.

	Each chapter builds upon previous material, with theorems and propositions proven rigorously. Algorithms are presented in pseudocode with complexity analysis. Applications are described with sufficient detail for implementation. The intended audience includes graduate students, researchers, and practitioners seeking deep understanding of attention mechanisms. While we maintain mathematical rigor, we also provide intuition and examples to make concepts accessible.
	
	\chapter{Mathematical Foundations}
	
	\section{Notation and Preliminaries}

	We establish notation that will be used throughout this monograph. Vectors are denoted by lowercase bold letters $\mathbf{v} \in \mathbb{R}^d$, matrices by uppercase bold letters $\mathbf{M} \in \mathbb{R}^{m \times n}$, and scalars by regular letters $x \in \mathbb{R}$. The $i$-th element of vector $\mathbf{v}$ is written $v_i$, and the element in row $i$ and column $j$ of matrix $\mathbf{M}$ is written $M_{ij}$ or $m_{ij}$. For a comprehensive reference of all notation and conventions, see \autoref{app:notation}.
	
	\subsection{Vector Spaces and Linear Transformations}
	
	Consider a vector space $\mathcal{V} = \mathbb{R}^d$ of dimension $d$. A linear transformation $T: \mathcal{V} \rightarrow \mathcal{W}$ from space $\mathcal{V}$ to space $\mathcal{W} = \mathbb{R}^{d'}$ can be represented by a matrix $\mathbf{W} \in \mathbb{R}^{d' \times d}$ such that:
	
	\begin{equation}
		T(\mathbf{v}) = \mathbf{W}\mathbf{v}
	\end{equation}
	
	The composition of two linear transformations $T_1: \mathcal{V}_1 \rightarrow \mathcal{V}_2$ and $T_2: \mathcal{V}_2 \rightarrow \mathcal{V}_3$ corresponds to matrix multiplication:
	
	\begin{equation}
		(T_2 \circ T_1)(\mathbf{v}) = \mathbf{W}_2(\mathbf{W}_1\mathbf{v}) = (\mathbf{W}_2\mathbf{W}_1)\mathbf{v}
	\end{equation}
	
	\subsection{Inner Products and Norms}
	
	The standard inner product on $\mathbb{R}^d$ is defined as:
	
	\begin{equation}
		\langle \mathbf{u}, \mathbf{v} \rangle = \mathbf{u}^T\mathbf{v} = \sum_{i=1}^{d} u_i v_i
	\end{equation}
	
	This inner product induces the Euclidean norm:
	
	\begin{equation}
		\|\mathbf{v}\|_2 = \sqrt{\langle \mathbf{v}, \mathbf{v} \rangle} = \sqrt{\sum_{i=1}^{d} v_i^2}
	\end{equation}
	
	The cosine similarity between two vectors measures the cosine of the angle between them:
	
	\begin{equation}
		\cos(\mathbf{u}, \mathbf{v}) = \frac{\langle \mathbf{u}, \mathbf{v} \rangle}{\|\mathbf{u}\|_2 \|\mathbf{v}\|_2}
	\end{equation}
	
	This similarity measure is bounded in $[-1, 1]$ and equals 1 when vectors are parallel, 0 when orthogonal, and -1 when anti-parallel.
	
	\subsection{Matrix Operations}
	
	The transpose of matrix $\mathbf{M} \in \mathbb{R}^{m \times n}$ is denoted $\mathbf{M}^T \in \mathbb{R}^{n \times m}$ with elements $(\mathbf{M}^T)_{ij} = M_{ji}$.
	
	The trace of a square matrix $\mathbf{M} \in \mathbb{R}^{n \times n}$ is the sum of diagonal elements:
	
	\begin{equation}
		\text{tr}(\mathbf{M}) = \sum_{i=1}^{n} M_{ii}
	\end{equation}
	
	The Frobenius norm of a matrix is:
	
	\begin{equation}
		\|\mathbf{M}\|_F = \sqrt{\sum_{i=1}^{m}\sum_{j=1}^{n} M_{ij}^2} = \sqrt{\text{tr}(\mathbf{M}^T\mathbf{M})}
	\end{equation}
	
	\section{Problem Formulation for Sequence Modeling}
	
	Consider an input sequence $\mathbf{X} = (\mathbf{x}_1, \mathbf{x}_2, \ldots, \mathbf{x}_n)$ where each $\mathbf{x}_i \in \mathbb{R}^{d_{in}}$ represents a vector at position $i$. In natural language processing, $\mathbf{x}_i$ might be a word embedding; in computer vision, it might be a patch embedding from an image.
	
	The sequence can be represented as a matrix $\mathbf{X} \in \mathbb{R}^{n \times d_{in}}$ where row $i$ contains vector $\mathbf{x}_i^T$. Our objective is to compute an output sequence $\mathbf{Y} = (\mathbf{y}_1, \mathbf{y}_2, \ldots, \mathbf{y}_n)$ where $\mathbf{y}_i \in \mathbb{R}^{d_{out}}$, such that each output vector aggregates information from the entire input sequence in a context-dependent manner.
	
	\subsection{Desiderata for Sequence Processing}
	
	An ideal sequence processing mechanism should satisfy several properties:
	
	\begin{enumerate}[leftmargin=*]
		\item \textbf{Flexibility}: The mechanism should allow each position to access any other position without architectural constraints.
		
		\item \textbf{Efficiency}: Computation should be parallelizable across positions to leverage modern hardware.
		
		\item \textbf{Adaptivity}: The patterns of information flow should be learned from data rather than hardcoded.
		
		\item \textbf{Bounded Memory}: Memory requirements should not grow excessively with sequence length.
		
		\item \textbf{Long-Range Modeling}: The mechanism should effectively capture dependencies between distant positions.
	\end{enumerate}
	
	Attention mechanisms satisfy most of these desiderata, though the memory requirement grows quadratically with sequence length in the standard formulation.
	
	\section{The General Attention Function}

	We now formally define the attention mechanism that forms the foundation for all subsequent developments. The notation introduced here for queries, keys, and values is used consistently throughout the monograph (see \autoref{tab:notation_attention} for a complete reference).
	
	\begin{definition}[Attention Mechanism]
		An attention mechanism is a differentiable function $\mathcal{A}: \mathbb{R}^{d_q} \times \mathbb{R}^{n \times d_k} \times \mathbb{R}^{n \times d_v} \rightarrow \mathbb{R}^{d_v}$ that computes a weighted combination of values based on the compatibility between a query and keys.
		
		Given:
		\begin{itemize}
			\item Query: $\mathbf{q} \in \mathbb{R}^{d_q}$
			\item Keys: $\mathbf{K} = (\mathbf{k}_1, \ldots, \mathbf{k}_n)$ where $\mathbf{k}_i \in \mathbb{R}^{d_k}$
			\item Values: $\mathbf{V} = (\mathbf{v}_1, \ldots, \mathbf{v}_n)$ where $\mathbf{v}_i \in \mathbb{R}^{d_v}$
		\end{itemize}
		
		The attention function computes:
		\begin{align}
			e_i &= \text{score}(\mathbf{q}, \mathbf{k}_i) \quad \text{for } i = 1, \ldots, n \label{eq:score}\\
			\alpha_i &= \frac{\exp(e_i)}{\sum_{j=1}^{n} \exp(e_j)} \label{eq:softmax}\\
			\mathcal{A}(\mathbf{q}, \mathbf{K}, \mathbf{V}) &= \sum_{i=1}^{n} \alpha_i \mathbf{v}_i \label{eq:weighted_sum}
		\end{align}
		
		where $\text{score}: \mathbb{R}^{d_q} \times \mathbb{R}^{d_k} \rightarrow \mathbb{R}$ is a compatibility function, and $\alpha_i$ are attention weights satisfying $\sum_{i=1}^n \alpha_i = 1$ and $\alpha_i \geq 0$.
	\end{definition}
	
	The attention mechanism consists of three conceptual stages:
	
	\textbf{Stage 1 - Score Computation}: For each position $i$, compute a scalar score $e_i$ measuring the compatibility or relevance between the query $\mathbf{q}$ and key $\mathbf{k}_i$. Different scoring functions are explored in the next section.
	
	\textbf{Stage 2 - Normalization}: Apply the softmax function to convert scores into a probability distribution. The softmax ensures that weights are non-negative and sum to one, allowing interpretation as a weighted average.
	
	\textbf{Stage 3 - Aggregation}: Compute the output as a weighted sum of value vectors, where each value is weighted by its corresponding attention weight.
	
	\begin{insight}
		The softmax normalization in Stage 2 is crucial for gradient flow during training. Alternative normalization schemes exist, but softmax provides a smooth, differentiable distribution that concentrates probability mass on high-scoring positions while maintaining non-zero weights for all positions.
	\end{insight}
	
	\subsection{Properties of the Attention Function}
	
	\begin{proposition}[Convex Combination]
		The output of the attention mechanism is a convex combination of the value vectors:
		\begin{equation}
			\mathcal{A}(\mathbf{q}, \mathbf{K}, \mathbf{V}) \in \text{conv}(\{\mathbf{v}_1, \ldots, \mathbf{v}_n\})
		\end{equation}
		where $\text{conv}$ denotes the convex hull.
	\end{proposition}
	
	\begin{trivlist}
		\item[\hskip\labelsep\bfseries Proof.]
		Since $\alpha_i \geq 0$ and $\sum_{i=1}^n \alpha_i = 1$, the output is by definition a convex combination of the value vectors.
	\end{trivlist}
	
	This property implies that the attention mechanism performs interpolation rather than extrapolation in the value space. The output cannot venture outside the region spanned by the input values.
	
	\begin{proposition}[Differentiability]
		The attention function $\mathcal{A}$ is differentiable with respect to all its inputs (query, keys, and values), provided the scoring function is differentiable.
	\end{proposition}
	
	\begin{trivlist}
		\item[\hskip\labelsep\bfseries Proof.]
		The softmax function is differentiable, and composition of differentiable functions is differentiable. The weighted sum is linear in the values and differentiable with respect to the weights. Therefore, the entire attention function is differentiable by the chain rule.
	\end{trivlist}
	
	This differentiability is essential for training neural networks via backpropagation.
	
	\section{Scoring Functions}
	
	The choice of scoring function significantly impacts the attention mechanism's computational properties and expressiveness. We analyze several commonly used scoring functions.
	
	\subsection{Additive (Bahdanau) Attention}

	The additive attention mechanism~\cite{bahdanau2014neural} computes scores using a feed-forward network:
	
	\begin{equation}
		\text{score}_{\text{add}}(\mathbf{q}, \mathbf{k}) = \mathbf{v}_a^T \tanh(\mathbf{W}_q\mathbf{q} + \mathbf{W}_k\mathbf{k})
	\end{equation}
	
	where $\mathbf{W}_q \in \mathbb{R}^{d_a \times d_q}$, $\mathbf{W}_k \in \mathbb{R}^{d_a \times d_k}$, and $\mathbf{v}_a \in \mathbb{R}^{d_a}$ are learned parameters, and $d_a$ is the attention hidden dimension.
	
	\textbf{Computational Complexity}: For $n$ keys and query dimension $d_q$, key dimension $d_k$:
	\begin{itemize}
		\item Computing $\mathbf{W}_q\mathbf{q}$: $O(d_a d_q)$
		\item Computing $\mathbf{W}_k\mathbf{k}_i$ for all $i$: $O(n d_a d_k)$
		\item Computing all scores: $O(n d_a)$
		\item Total: $O(n d_a (d_k + 1) + d_a d_q)$
	\end{itemize}
	
	\textbf{Properties}:
	\begin{itemize}
		\item Can handle queries and keys of different dimensions
		\item Requires learning $O(d_a(d_q + d_k + 1))$ parameters
		\item The $\tanh$ nonlinearity provides additional expressiveness
	\end{itemize}
	
	\subsection{Multiplicative (Luong) Attention}

	The multiplicative attention~\cite{luong2015effective} computes scores via a bilinear form:
	
	\begin{equation}
		\text{score}_{\text{mult}}(\mathbf{q}, \mathbf{k}) = \mathbf{q}^T\mathbf{W}\mathbf{k}
	\end{equation}
	
	where $\mathbf{W} \in \mathbb{R}^{d_q \times d_k}$ is a learned weight matrix.
	
	\textbf{Computational Complexity}:
	\begin{itemize}
		\item Computing $\mathbf{W}\mathbf{k}_i$ for all $i$: $O(n d_q d_k)$
		\item Computing all $\mathbf{q}^T(\mathbf{W}\mathbf{k}_i)$: $O(n d_q)$
		\item Total: $O(n d_q d_k)$
	\end{itemize}
	
	\textbf{Properties}:
	\begin{itemize}
		\item More efficient than additive attention when $d_a > \max(d_q, d_k)$
		\item Requires learning $O(d_q d_k)$ parameters
		\item Can handle different query and key dimensions
	\end{itemize}
	
	\subsection{Dot-Product Attention}
	
	When query and key dimensions match ($d_q = d_k = d$), we can use simple dot-product:
	
	\begin{equation}
		\text{score}_{\text{dot}}(\mathbf{q}, \mathbf{k}) = \mathbf{q}^T\mathbf{k} = \sum_{i=1}^{d} q_i k_i
	\end{equation}
	
	\textbf{Computational Complexity}:
	\begin{itemize}
		\item Computing all scores: $O(nd)$
	\end{itemize}
	
	\textbf{Properties}:
	\begin{itemize}
		\item No learnable parameters in the scoring function itself
		\item Maximum computational efficiency
		\item Requires $d_q = d_k$
	\end{itemize}
	
	The dot product measures similarity between query and key vectors. When both are unit vectors, the dot product equals the cosine of the angle between them.
	
	\subsection{Scaled Dot-Product Attention}
	
	To address numerical stability issues with dot-product attention, the Transformer architecture~\cite{vaswani2017attention} introduced scaling:
	
	\begin{equation}
		\text{score}_{\text{scaled}}(\mathbf{q}, \mathbf{k}) = \frac{\mathbf{q}^T\mathbf{k}}{\sqrt{d_k}}
	\end{equation}
	
	\begin{theorem}[Variance of Dot Product]
		Assume query and key components are independent random variables with mean 0 and variance 1. Then the dot product $\mathbf{q}^T\mathbf{k}$ has mean 0 and variance $d_k$.
	\end{theorem}
	
	\begin{trivlist}
		\item[\hskip\labelsep\bfseries Proof.]
		Let $\mathbf{q}, \mathbf{k} \in \mathbb{R}^{d_k}$ with components $q_i, k_i$ independently distributed with $\mathbb{E}[q_i] = \mathbb{E}[k_i] = 0$ and $\text{Var}(q_i) = \text{Var}(k_i) = 1$.
		
		The dot product is:
		\begin{equation}
			\mathbf{q}^T\mathbf{k} = \sum_{i=1}^{d_k} q_i k_i
		\end{equation}
		
		The expected value is:
		\begin{equation}
			\mathbb{E}[\mathbf{q}^T\mathbf{k}] = \sum_{i=1}^{d_k} \mathbb{E}[q_i k_i] = \sum_{i=1}^{d_k} \mathbb{E}[q_i]\mathbb{E}[k_i] = 0
		\end{equation}
		
		The variance is:
		\begin{align}
			\text{Var}(\mathbf{q}^T\mathbf{k}) &= \mathbb{E}[(\mathbf{q}^T\mathbf{k})^2] - (\mathbb{E}[\mathbf{q}^T\mathbf{k}])^2 \\
			&= \mathbb{E}\left[\sum_{i=1}^{d_k}\sum_{j=1}^{d_k} q_i k_i q_j k_j\right] \\
			&= \sum_{i=1}^{d_k}\sum_{j=1}^{d_k} \mathbb{E}[q_i q_j k_i k_j]
		\end{align}
		
		For $i \neq j$, by independence:
		\begin{equation}
			\mathbb{E}[q_i q_j k_i k_j] = \mathbb{E}[q_i]\mathbb{E}[q_j]\mathbb{E}[k_i]\mathbb{E}[k_j] = 0
		\end{equation}
		
		For $i = j$:
		\begin{equation}
			\mathbb{E}[q_i^2 k_i^2] = \mathbb{E}[q_i^2]\mathbb{E}[k_i^2] = 1 \cdot 1 = 1
		\end{equation}
		
		Therefore:
		\begin{equation}
			\text{Var}(\mathbf{q}^T\mathbf{k}) = \sum_{i=1}^{d_k} 1 = d_k
		\end{equation}
	\end{trivlist}
	
	\begin{corollary}[Variance of Scaled Dot Product]
		The scaled dot product $\frac{\mathbf{q}^T\mathbf{k}}{\sqrt{d_k}}$ has variance 1 under the same assumptions.
	\end{corollary}
	
	\begin{keypoint}
		Scaling by $1/\sqrt{d_k}$ maintains unit variance in the scores regardless of dimension, preventing the softmax from entering regions with vanishingly small gradients when dimensions are large.
	\end{keypoint}
	
	When scores have a large magnitude, the softmax concentrates almost all probability mass on the maximum score, with gradients approaching zero for all positions except the maximum. Scaling prevents this saturation.
	
	\section{Vectorized Formulation}
	
	For computational efficiency, attention is typically computed for multiple queries simultaneously using matrix operations. This vectorization enables exploitation of hardware parallelism and optimized linear algebra libraries.
	
	\subsection{Multiple Queries}
	
	Let $\mathbf{Q} \in \mathbb{R}^{m \times d_k}$ represent $m$ query vectors (rows of $\mathbf{Q}$), $\mathbf{K} \in \mathbb{R}^{n \times d_k}$ represent $n$ key vectors, and $\mathbf{V} \in \mathbb{R}^{n \times d_v}$ represent $n$ value vectors.
	
	The attention operation for all queries can be computed as:
	
	\begin{equation}
		\text{Attention}(\mathbf{Q}, \mathbf{K}, \mathbf{V}) = \text{softmax}\left(\frac{\mathbf{Q}\mathbf{K}^T}{\sqrt{d_k}}\right)\mathbf{V}
	\end{equation}
	
	Let us decompose this operation:
	
	\textbf{Step 1 - Score Matrix}: Compute the matrix product $\mathbf{S} = \mathbf{Q}\mathbf{K}^T \in \mathbb{R}^{m \times n}$. Element $S_{ij}$ equals the dot product between query $i$ and key $j$:
	\begin{equation}
		S_{ij} = \mathbf{q}_i^T \mathbf{k}_j = \sum_{\ell=1}^{d_k} q_{i\ell} k_{j\ell}
	\end{equation}
	
	\textbf{Step 2 - Scaling}: Compute $\tilde{\mathbf{S}} = \mathbf{S}/\sqrt{d_k}$.
	
	\textbf{Step 3 - Attention Weights}: Apply softmax row-wise to produce the attention weight matrix $\mathbf{A} \in \mathbb{R}^{m \times n}$:
	\begin{equation}
		A_{ij} = \frac{\exp(\tilde{S}_{ij})}{\sum_{k=1}^{n} \exp(\tilde{S}_{ik})}
	\end{equation}
	
	Each row of $\mathbf{A}$ forms a probability distribution over the $n$ keys.
	
	\textbf{Step 4 - Weighted Sum}: Compute the output $\mathbf{O} = \mathbf{A}\mathbf{V} \in \mathbb{R}^{m \times d_v}$. Row $i$ of $\mathbf{O}$ is:
	\begin{equation}
		\mathbf{o}_i = \sum_{j=1}^{n} A_{ij} \mathbf{v}_j
	\end{equation}

\begin{figure}[t]
		\centering
		\begin{tikzpicture}[scale=0.95]
			\node[draw, rectangle, minimum width=2.2cm, minimum height=1.2cm, 
			fill=anthropicLightOrange!40, thick] (Q) at (0, 5) {$\mathbf{Q}$};
			\node[below=0.1cm of Q, font=\small] {$m \times d_k$};
			
			\node[draw, rectangle, minimum width=2.2cm, minimum height=1.2cm, 
			fill=anthropicBeige!60, thick] (K) at (3.5, 5) {$\mathbf{K}$};
			\node[below=0.1cm of K, font=\small] {$n \times d_k$};
			
			\node[draw, rectangle, minimum width=2.2cm, minimum height=1.2cm, 
			fill=anthropicTan!50, thick] (V) at (7, 5) {$\mathbf{V}$};
			\node[font=\small, right=-1.1cm of V.south east, yshift=-0.3cm] {$n \times d_v$};
			
			\node[draw, rectangle, minimum width=3cm, minimum height=1cm, 
			fill=anthropicOrange!30, thick] (QKT) at (1.75, 3) {$\mathbf{Q}\mathbf{K}^T$};
			\node[font=\small, right=-1.3cm of QKT.south east, yshift=-0.2cm] {$m \times n$};
			\draw[-Stealth, very thick, anthropicBrown] (Q) -- (QKT);
			\draw[-Stealth, very thick, anthropicBrown] (K) -- (QKT);
			
			\node[draw, rectangle, minimum width=3cm, minimum height=0.8cm, 
			fill=anthropicCream, thick] (scale) at (1.75, 1.5) {Scale: $\div\sqrt{d_k}$};
			\draw[-Stealth, very thick, anthropicBrown] (QKT) -- (scale);
			
			\node[draw, rectangle, minimum width=3cm, minimum height=0.8cm, 
			fill=anthropicOrange!50, thick] (soft) at (1.75, 0.2) {Softmax (row-wise)};
			\draw[-Stealth, very thick, anthropicBrown] (scale) -- (soft);
			
			\node[draw, rectangle, minimum width=2.5cm, minimum height=1cm, 
			fill=anthropicDeepOrange!30, thick] (A) at (1.75, -1.3) {$\mathbf{A}$};
			\node[below=0.1cm of A, font=\small] {$m \times n$};
			\draw[-Stealth, very thick, anthropicBrown] (soft) -- (A);
			
			\node[draw, rectangle, minimum width=2.5cm, minimum height=1.2cm, 
			fill=anthropicLightOrange!60, thick] (output) at (7, -1.3) {Output};
			\node[below=0.1cm of output, font=\small] {$m \times d_v$};
			\draw[-Stealth, very thick, anthropicBrown] (A) -- (output);
			\draw[-Stealth, very thick, anthropicBrown] (V) to[out=270, in=90] (output);
			
			\node[align=left, font=\small, text=anthropicBrown] at (9.5, 1.5) {
				\textbf{Dimensions:}\\
				$m$: \# queries\\
				$n$: \# keys/values\\
				$d_k$: key dimension\\
				$d_v$: value dimension\\[0.3em]
				\textbf{Complexity:}\\
				$O(mn(d_k + d_v))$
			};
		\end{tikzpicture}
		\caption{Computational flow of vectorized scaled dot-product attention. The attention mechanism computes similarities between all query-key pairs, normalizes them via softmax, and uses the resulting weights to aggregate values.}
		\label{fig:attention_flow_detailed}
	\end{figure}
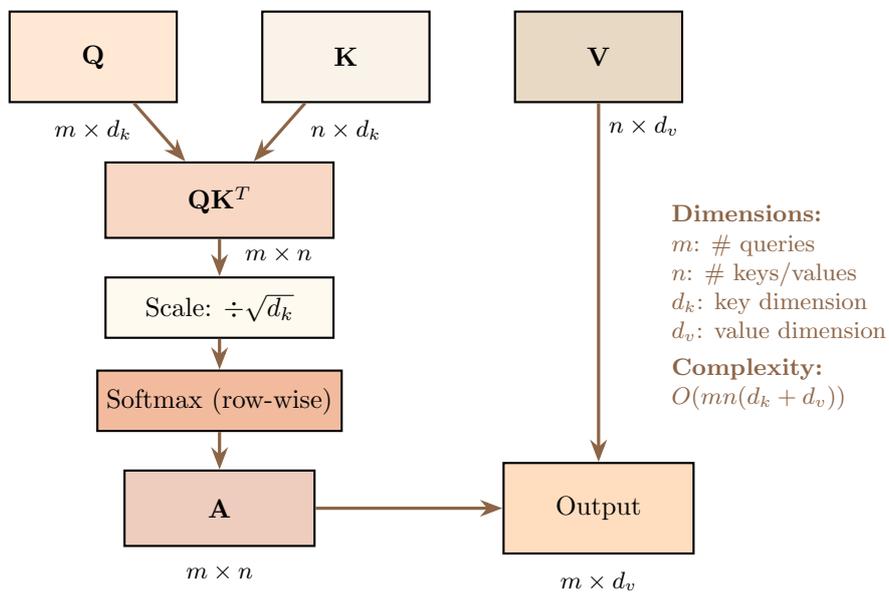

	The complete computational flow is illustrated in \autoref{fig:attention_flow_detailed}.

	\subsection{Computational Complexity}
	
	Let us analyze the computational complexity of each step:
	
	\begin{enumerate}[leftmargin=*]
		\item \textbf{Score Matrix} $\mathbf{Q}\mathbf{K}^T$: $O(m \cdot n \cdot d_k)$ operations
		\item \textbf{Scaling}: $O(m \cdot n)$ operations (negligible)
		\item \textbf{Softmax}: $O(m \cdot n)$ operations for exponentials and normalization
		\item \textbf{Weighted Sum} $\mathbf{A}\mathbf{V}$: $O(m \cdot n \cdot d_v)$ operations
	\end{enumerate}
	
	Total complexity: $O(m \cdot n \cdot (d_k + d_v))$
	
	When $m = n$ (self-attention case) and $d_k = d_v = d$:
	\begin{equation}
		\text{Complexity} = O(n^2 d)
	\end{equation}
	
	This quadratic dependence on sequence length is the primary computational limitation of attention mechanisms for long sequences.
	
	\subsection{Memory Requirements}
	
	The attention mechanism requires storing:
	\begin{itemize}
		\item Score matrix $\mathbf{S}$: $O(mn)$ memory
		\item Attention weights $\mathbf{A}$: $O(mn)$ memory
		\item Intermediate gradients during backpropagation: $O(mn)$ memory
	\end{itemize}
	
	Total memory: $O(mn)$, or $O(n^2)$ for self-attention.
	
	This quadratic memory requirement limits the maximum sequence length that can be processed on available hardware.
	
	\subsection{Implementation Considerations}
	
	Efficient implementation of attention requires careful consideration of numerical stability and computational efficiency:
	
	\textbf{Numerical Stability in Softmax}: Computing $\exp(x)$ for large $x$ can cause overflow. The numerically stable implementation subtracts the maximum from all values:
	
	\begin{equation}
		\text{softmax}(\mathbf{x})_i = \frac{\exp(x_i - \max_j x_j)}{\sum_{j=1}^{n} \exp(x_j - \max_j x_j)}
	\end{equation}
	
	This transformation does not change the result, but prevents overflow.
	
	\textbf{Fused Operations}: Modern deep learning frameworks fuse the scaling and softmax operations for efficiency, computing:
	
	\begin{equation}
		\text{softmax}\left(\frac{\mathbf{Q}\mathbf{K}^T}{\sqrt{d_k}}\right)
	\end{equation}
	
	in a single kernel, reducing memory transfers.
	
	\textbf{Mixed Precision}: Using 16-bit floating point for matrix multiplications while maintaining 32-bit precision for softmax calculations balances speed and numerical stability.
	
	\chapter{Self-Attention Mechanisms}
	
	Self-attention represents a fundamental innovation in which a sequence attends to itself, allowing each position to gather information from all positions in the same sequence. This mechanism forms the core of the Transformer architecture and enables modeling of complex intra-sequence dependencies.
	
	\section{Definition and Formulation}
	
	Unlike traditional attention mechanisms that operate between distinct source and target sequences (e.g., encoder and decoder in machine translation), self-attention computes attention within a single sequence.
	
	\begin{definition}[Self-Attention]
		Given an input sequence represented as a matrix $\mathbf{X} \in \mathbb{R}^{n \times d_{model}}$ where $n$ is the sequence length and $d_{model}$ is the model dimension, self-attention computes queries, keys, and values through learned linear projections:
		
		\begin{align}
			\mathbf{Q} &= \mathbf{X}\mathbf{W}^Q \label{eq:self_attn_q}\\
			\mathbf{K} &= \mathbf{X}\mathbf{W}^K \label{eq:self_attn_k}\\
			\mathbf{V} &= \mathbf{X}\mathbf{W}^V \label{eq:self_attn_v}
		\end{align}
		
		where $\mathbf{W}^Q, \mathbf{W}^K \in \mathbb{R}^{d_{model} \times d_k}$ and $\mathbf{W}^V \in \mathbb{R}^{d_{model} \times d_v}$ are learned parameter matrices.
		
		The self-attention operation then computes:
		\begin{equation}
			\text{SelfAttention}(\mathbf{X}) = \text{softmax}\left(\frac{\mathbf{Q}\mathbf{K}^T}{\sqrt{d_k}}\right)\mathbf{V}
		\end{equation}
	\end{definition}
	
	The output is a matrix $\mathbf{Y} \in \mathbb{R}^{n \times d_v}$ where row $i$ contains a contextualized representation of position $i$, aggregating information from all positions according to learned attention weights.
	
	\subsection{Interpretation}
	
	Each position $i$ in the sequence generates:
	\begin{itemize}
		\item A \textbf{query} $\mathbf{q}_i = \mathbf{x}_i \mathbf{W}^Q$ representing what information it seeks
		\item A \textbf{key} $\mathbf{k}_i = \mathbf{x}_i \mathbf{W}^K$ representing what information it offers
		\item A \textbf{value} $\mathbf{v}_i = \mathbf{x}_i \mathbf{W}^V$ representing the actual information content
	\end{itemize}
	
	The attention weights $\alpha_{ij}$ determine how much position $i$ attends to position $j$:
	
	\begin{equation}
		\alpha_{ij} = \frac{\exp\left(\frac{\mathbf{q}_i^T \mathbf{k}_j}{\sqrt{d_k}}\right)}{\sum_{k=1}^{n} \exp\left(\frac{\mathbf{q}_i^T \mathbf{k}_k}{\sqrt{d_k}}\right)}
	\end{equation}
	
	The output at position $i$ is then:
	\begin{equation}
		\mathbf{y}_i = \sum_{j=1}^{n} \alpha_{ij} \mathbf{v}_j
	\end{equation}
	
	This formulation allows the network to learn what patterns of attention are useful for the task at hand, rather than imposing predefined connectivity structures.
	
	\section{Mathematical Properties}
	
	Self-attention exhibits several important mathematical properties that distinguish it from other sequence modeling approaches.
	
	\subsection{Permutation Equivariance}
	
	\begin{theorem}[Permutation Equivariance]
		Self-attention without positional information is equivariant to permutations of the input sequence. That is, for any permutation matrix $\mathbf{P} \in \{0,1\}^{n \times n}$ (where exactly one entry in each row and column is 1):
		
		\begin{equation}
			\text{SelfAttention}(\mathbf{P}\mathbf{X}) = \mathbf{P} \cdot \text{SelfAttention}(\mathbf{X})
		\end{equation}
	\end{theorem}
	
	\begin{trivlist}
		\item[\hskip\labelsep\bfseries Proof.]
		Let $\mathbf{Y} = \text{SelfAttention}(\mathbf{X})$. We need to show that permuting the input by $\mathbf{P}$ results in the same permutation of the output.
		
		For input $\mathbf{X}' = \mathbf{P}\mathbf{X}$:
		\begin{align}
			\mathbf{Q}' &= \mathbf{X}'\mathbf{W}^Q = \mathbf{P}\mathbf{X}\mathbf{W}^Q = \mathbf{P}\mathbf{Q} \\
			\mathbf{K}' &= \mathbf{X}'\mathbf{W}^K = \mathbf{P}\mathbf{X}\mathbf{W}^K = \mathbf{P}\mathbf{K} \\
			\mathbf{V}' &= \mathbf{X}'\mathbf{W}^V = \mathbf{P}\mathbf{X}\mathbf{W}^V = \mathbf{P}\mathbf{V}
		\end{align}
		
		The score matrix becomes:
		\begin{equation}
			\mathbf{S}' = \mathbf{Q}'\mathbf{K}'^T = (\mathbf{P}\mathbf{Q})(\mathbf{P}\mathbf{K})^T = \mathbf{P}\mathbf{Q}\mathbf{K}^T\mathbf{P}^T = \mathbf{P}\mathbf{S}\mathbf{P}^T
		\end{equation}
		
		Since $\mathbf{P}$ is a permutation matrix, $\mathbf{P}\mathbf{S}\mathbf{P}^T$ permutes both rows and columns of $\mathbf{S}$ according to $\mathbf{P}$.
		
		The attention weights matrix:
		\begin{equation}
			\mathbf{A}' = \text{softmax}(\mathbf{S}') = \text{softmax}(\mathbf{P}\mathbf{S}\mathbf{P}^T)
		\end{equation}
		
		Since softmax is applied row-wise and permutation preserves row structure:
		\begin{equation}
			\mathbf{A}' = \mathbf{P} \cdot \text{softmax}(\mathbf{S}) \cdot \mathbf{P}^T = \mathbf{P}\mathbf{A}\mathbf{P}^T
		\end{equation}
		
		Finally, the output:
		\begin{equation}
			\mathbf{Y}' = \mathbf{A}'\mathbf{V}' = (\mathbf{P}\mathbf{A}\mathbf{P}^T)(\mathbf{P}\mathbf{V}) = \mathbf{P}\mathbf{A}\mathbf{V} = \mathbf{P}\mathbf{Y}
		\end{equation}
		
		Therefore, $\text{SelfAttention}(\mathbf{P}\mathbf{X}) = \mathbf{P} \cdot \text{SelfAttention}(\mathbf{X})$.
	\end{trivlist}
	
	\begin{trivlist}
		\item[\hskip\labelsep\bfseries Corollary (Order Invariance).]
		Without positional information, self-attention is a set function rather than a sequence function. The output set is independent of input ordering, though the correspondence between input and output positions is preserved.
	\end{trivlist}
		
	This property has profound implications: self-attention has no inherent notion of sequence order. To process sequences where order matters (such as text), explicit positional information must be injected, typically through positional encodings added to the input embeddings.
	
	\subsection{Computational Complexity}
	
	\begin{proposition}[Time Complexity]
		For a sequence of length $n$ with model dimension $d_{model}$ and attention dimension $d_k = d_v = d$, self-attention has time complexity:
		\begin{equation}
			T(n) = O(n^2 d + n d_{model}^2)
		\end{equation}
	\end{proposition}
	
	\begin{trivlist}
		\item[\hskip\labelsep\bfseries Proof.]
		The computation involves:
		\begin{enumerate}
			\item Computing $\mathbf{Q}$, $\mathbf{K}$, $\mathbf{V}$: Each is $\mathbf{X}\mathbf{W}$ where $\mathbf{X} \in \mathbb{R}^{n \times d_{\text{model}}}$ and $\mathbf{W} \in \mathbb{R}^{d_{\text{model}} \times d}$. Cost: $3 \times O(n d_{\text{model}} d) = O(n d_{\text{model}} d)$.
			
			\item Computing $\mathbf{Q}\mathbf{K}^T$: Matrix multiplication of $\mathbf{Q} \in \mathbb{R}^{n \times d}$ and $\mathbf{K}^T \in \mathbb{R}^{d \times n}$. Cost: $O(n^2 d)$.
			
			\item Computing softmax: Row-wise operation on $n \times n$ matrix. Cost: $O(n^2)$.
			
			\item Computing $\mathbf{A}\mathbf{V}$: Matrix multiplication of $\mathbf{A} \in \mathbb{R}^{n \times n}$ and $\mathbf{V} \in \mathbb{R}^{n \times d}$. Cost: $O(n^2 d)$.
		\end{enumerate}
		
		Typically, $d_{\text{model}} \geq d$ (often $d = d_{\text{model}}$ or $d = d_{\text{model}}/h$ for $h$ heads). When $d = d_{\text{model}}$, the dominant terms are:
		\begin{equation}
			O(n d_{\text{model}}^2) + O(n^2 d_{\text{model}})
		\end{equation}
		
		For long sequences where $n > d_{\text{model}}$, the $O(n^2 d_{\text{model}})$ term dominates. For short sequences where $n < d_{\text{model}}$, the $O(n d_{\text{model}}^2)$ term dominates.
	\end{trivlist}

	\begin{proposition}[Space Complexity]
		Self-attention requires $O(n^2 + n d_{model})$ memory for activations and intermediate values.
	\end{proposition}
	
	\begin{trivlist}
		\item[\hskip\labelsep\bfseries Proof.]
		Memory requirements:
		\begin{itemize}
			\item Input $\mathbf{X}$: $O(n d_{\text{model}})$
			\item $\mathbf{Q}$, $\mathbf{K}$, $\mathbf{V}$: $3 \times O(nd) = O(nd)$
			\item Score matrix $\mathbf{S}$ and attention weights $\mathbf{A}$: $2 \times O(n^2) = O(n^2)$
			\item Output $\mathbf{Y}$: $O(nd)$
		\end{itemize}
		
		Total: $O(n^2 + nd + n d_{\text{model}})$. For typical cases where $d \approx d_{\text{model}}$, this is $O(n^2 + n d_{\text{model}})$.
		
		The $O(n^2)$ term is the primary memory bottleneck for long sequences.
	\end{trivlist}

	\subsection{Comparison with Recurrent Networks}
	
	\begin{table}[h]
		\centering
		\caption{Comparison of sequence modeling architectures}
		\begin{tabular}{lccc}
			\toprule
			\textbf{Architecture} & \textbf{Complexity} & \textbf{Sequential} & \textbf{Path Length} \\
			\midrule
			Self-Attention & $O(n^2 \cdot d)$ & $O(1)$ & $O(1)$ \\
			Recurrent (RNN/LSTM) & $O(n \cdot d^2)$ & $O(n)$ & $O(n)$ \\
			Convolutional & $O(k \cdot n \cdot d^2)$ & $O(1)$ & $O(\log_k(n))$ \\
			\bottomrule
		\end{tabular}
		\label{tab:architecture_comparison}
	\end{table}
	
	\autoref{tab:architecture_comparison} compares three sequence modeling approaches:
	
	\textbf{Complexity}: Per-layer computational cost for sequence length $n$ and dimension $d$. Self-attention is quadratic in $n$, while recurrent networks are linear in $n$ but quadratic in $d$.
	
	\textbf{Sequential Operations}: Minimum number of sequential operations required, affecting parallelizability. Self-attention and convolutions allow parallel computation, while RNNs require sequential processing.
	
	\textbf{Maximum Path Length}: Maximum number of operations between any two positions in the sequence. Self-attention allows direct connections ($O(1)$), while RNNs require traversing all intermediate positions ($O(n)$), and convolutions require logarithmic depth for full receptive field.
	
	The $O(1)$ path length of self-attention is crucial for learning long-range dependencies, as gradients can flow directly between distant positions without being attenuated through intermediate states.
	
	\section{Attention Patterns}
	
	The attention weight matrix $\mathbf{A} \in \mathbb{R}^{n \times n}$ captures pairwise relationships between all positions. Entry $A_{ij}$ indicates how much position $i$ attends to position $j$. Empirical analysis of trained models reveals diverse learned patterns.
	
	\subsection{Pattern Taxonomy}
	
	\begin{example}[Local Patterns]
		In early layers of language models, attention weights often concentrate near the diagonal, with high values $A_{ij}$ when $|i - j|$ is small. This pattern captures local dependencies similar to n-grams in traditional NLP.
		
		The attention distribution for position $i$ approximates:
		\begin{equation}
			A_{ij} \approx \frac{1}{Z_i} \exp\left(-\frac{|i-j|^2}{2\sigma^2}\right)
		\end{equation}
		where $Z_i$ is a normalization constant and $\sigma$ controls the window width.
	\end{example}
	
	\begin{example}[Positional Patterns]
		Some attention heads learn to attend to specific relative positions. For instance, a head might attend primarily to the next position ($A_{i,i+1} \approx 1$) or the previous position ($A_{i,i-1} \approx 1$). These patterns extract positional information.
	\end{example}
	
	\begin{example}[Syntactic Patterns]
		In language models, certain heads learn attention patterns that follow syntactic structure. For example:
		\begin{itemize}
			\item Heads that attend from verbs to their subjects
			\item Heads that attend from pronouns to their antecedents
			\item Heads that attend from words to their modifiers
		\end{itemize}
		
		These patterns emerge without explicit syntactic supervision, suggesting that attention mechanisms discover grammatical structure through language modeling objectives.
	\end{example}
	
	\begin{example}[Broadcast Patterns]
		Some heads develop attention patterns where certain positions (often the first position or special tokens) receive attention from most other positions. These positions act as information aggregators, collecting global information that is then broadcast to other positions in subsequent layers.
	\end{example}
	
	\subsection{Attention Entropy}
	
	The entropy of attention distribution for position $i$ measures its focus:
	
	\begin{equation}
		H_i = -\sum_{j=1}^{n} A_{ij} \log A_{ij}
	\end{equation}
	
	\begin{itemize}
		\item \textbf{Low entropy} ($H_i \approx 0$): Attention focused on few positions
		\item \textbf{High entropy} ($H_i \approx \log n$): Attention distributed broadly
	\end{itemize}
	
	Empirical studies show that:
	\begin{itemize}
		\item Early layers tend to have lower entropy (more focused attention)
		\item Later layers often have higher entropy (more distributed attention)
		\item Different heads within the same layer exhibit different entropy profiles
	\end{itemize}
	
	\section{Masked Self-Attention}
	
	For autoregressive modeling (e.g., language modeling where each position should only depend on previous positions), we use masked self-attention.
	
	\begin{definition}[Causal Masking]
		In masked self-attention, position $i$ is prevented from attending to positions $j > i$ by setting:
		
		\begin{equation}
			\tilde{S}_{ij} = \begin{cases}
				S_{ij} & \text{if } j \leq i \\
				-\infty & \text{if } j > i
			\end{cases}
		\end{equation}
		
		before applying softmax. Since $\exp(-\infty) = 0$, this ensures $A_{ij} = 0$ for $j > i$.
	\end{definition}
	
	In practice, we use a large negative number (e.g., $-10^{10}$) rather than $-\infty$ to avoid numerical issues.
	
	\begin{proposition}[Causality Property]
		Masked self-attention ensures that the output at position $i$ depends only on inputs at positions $1, \ldots, i$:
		\begin{equation}
			\mathbf{y}_i = f(\mathbf{x}_1, \ldots, \mathbf{x}_i)
		\end{equation}
	\end{proposition}
	
	This property is essential for autoregressive generation, where we generate sequence elements one at a time, each conditioned only on previous elements.

\begin{figure}[t]
		\centering
		\begin{tikzpicture}[scale=0.8]
			\node[font=\bfseries, color=anthropicDeepOrange] at (-1.1, 4.7) {Unmasked};
			\foreach \i in {1,...,5} {
				\foreach \j in {1,...,5} {
					\pgfmathsetmacro{\opacity}{0.2 + 0.6*rnd}
					\fill[anthropicOrange, opacity=\opacity] 
					(-4+0.8*\j, 3-0.8*\i) rectangle (-3.2+0.8*\j, 3.8-0.8*\i);
				}
			}
			
			\node[font=\small] at (-3.6, 2.6) {1};
			\node[font=\small] at (-3.6, 1.8) {2};
			\node[font=\small] at (-3.6, 1.0) {3};
			\node[font=\small] at (-3.6, 0.2) {4};
			\node[font=\small] at (-3.6, -0.6) {5};
			
			\node[font=\small] at (-2.8, 3.2) {1};
			\node[font=\small] at (-2.0, 3.2) {2};
			\node[font=\small] at (-1.2, 3.2) {3};
			\node[font=\small] at (-0.4, 3.2) {4};
			\node[font=\small] at (0.4, 3.2) {5};
			
			\node[font=\bfseries, color=anthropicDeepOrange] at (4.7, 4.7) {Masked (Causal)};
			\foreach \i in {1,...,5} {
				\foreach \j in {1,...,\i} {
					\pgfmathsetmacro{\opacity}{0.2 + 0.6*rnd}
					\fill[anthropicOrange, opacity=\opacity] 
					(2+0.8*\j, 3-0.8*\i) rectangle (2.8+0.8*\j, 3.8-0.8*\i);
				}
				\foreach \j in {\i,...,5} {
					\fill[anthropicBeige] 
					(2.02+0.8*\j, 3.02-0.8*\i) rectangle (2.78+0.8*\j, 3.78-0.8*\i);
					\draw[anthropicBrown, very thin] 
					(2.1+0.8*\j, 3.1-0.8*\i) -- (2.7+0.8*\j, 3.7-0.8*\i);
					\draw[anthropicBrown, very thin] 
					(2.7+0.8*\j, 3.1-0.8*\i) -- (2.1+0.8*\j, 3.7-0.8*\i);
				}
			}
			
			\node[font=\small] at (2.4, 2.6) {1};
			\node[font=\small] at (2.4, 1.8) {2};
			\node[font=\small] at (2.4, 1.0) {3};
			\node[font=\small] at (2.4, 0.2) {4};
			\node[font=\small] at (2.4, -0.6) {5};
			
			\node[font=\small] at (3.2, 3.2) {1};
			\node[font=\small] at (4.0, 3.2) {2};
			\node[font=\small] at (4.8, 3.2) {3};
			\node[font=\small] at (5.6, 3.2) {4};
			\node[font=\small] at (6.4, 3.2) {5};
			
			\node[font=\small, align=center] at (-1, -1.5) {Query\\Position};
			\node[font=\small] at (-1, 3.9) {Key Position};
			\node[font=\small, align=center] at (5, -1.5) {Query\\Position};
			\node[font=\small] at (5, 3.9) {Key Position};
			
			\fill[anthropicOrange, opacity=0.6] (7.6, 2.5) rectangle (8.1, 3);
			\node[right, font=\small] at (8.3, 2.75) {Attention weight};
			\fill[anthropicBeige] (7.6, 1.8) rectangle (8.1, 2.3);
			\draw[anthropicBrown, very thin] (7.6, 1.9) -- (8.1, 2.2);
			\draw[anthropicBrown, very thin] (8.1, 1.9) -- (7.6, 2.2);
			\node[right, font=\small] at (8.3, 2.05) {Masked (no attention)};
		\end{tikzpicture}
		\caption{Visualization of attention weight matrices for unmasked (left) and masked/causal (right) self-attention. In masked attention, positions cannot attend to future positions, enforcing temporal causality.}
		\label{fig:masked_attention}
	\end{figure}
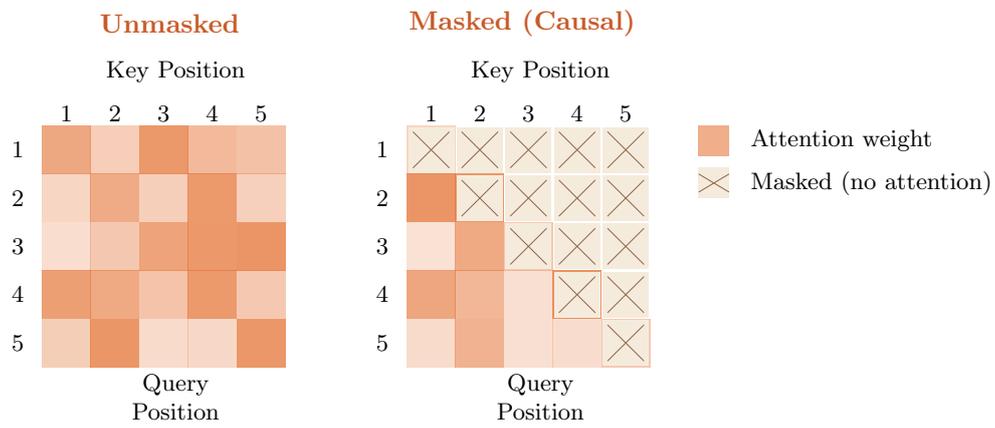

	The masking pattern forms a lower triangular attention matrix, where $A_{ij} > 0$ only when $j \leq i$. This allows parallel training on complete sequences while maintaining the causal structure needed for generation. \autoref{fig:masked_attention} illustrates the difference between unmasked and masked (causal) self-attention patterns.
	
	\chapter{Multi-Head Attention}
	
	A single attention mechanism may be limited in its capacity to capture different types of relationships simultaneously. Multi-head attention addresses this limitation by computing multiple attention functions in parallel, each with independent parameters, allowing the model to attend to information from different representation subspaces at different positions.
	
	\section{Motivation and Architecture}
	
	Consider the task of understanding the sentence ``The animal didn't cross the street because it was too tired." The pronoun ``it" could potentially refer to either ``animal" or ``street," though in this context it refers to ``animal." Determining this requires understanding:
	
	\begin{enumerate}[leftmargin=*]
		\item Syntactic structure (subject-verb relationships)
		\item Semantic plausibility (animals get tired, streets don't)
		\item Long-range dependencies (the pronoun appears far from its antecedent)
	\end{enumerate}
	
	A single attention mechanism must balance these different types of information. Multi-head attention allows different heads to specialize in different aspects, some focusing on syntax, others on semantics, others on positional patterns.
	
	\subsection{Formal Definition}
	
	\begin{definition}[Multi-Head Attention]
		Multi-head attention with $h$ heads computes $h$ parallel attention functions, each with separate parameters, then concatenates and linearly projects the results.
		
		Given input matrices $\mathbf{Q}, \mathbf{K}, \mathbf{V} \in \mathbb{R}^{n \times d_{model}}$:
		
		\begin{align}
			\text{head}_i &= \text{Attention}(\mathbf{Q}\mathbf{W}_i^Q, \mathbf{K}\mathbf{W}_i^K, \mathbf{V}\mathbf{W}_i^V) \label{eq:mha_head}\\
			\text{MultiHead}(\mathbf{Q}, \mathbf{K}, \mathbf{V}) &= \text{Concat}(\text{head}_1, \ldots, \text{head}_h)\mathbf{W}^O \label{eq:mha_concat}
		\end{align}
		
		where:
		\begin{itemize}
			\item $\mathbf{W}_i^Q, \mathbf{W}_i^K \in \mathbb{R}^{d_{model} \times d_k}$: Query and key projections for head $i$
			\item $\mathbf{W}_i^V \in \mathbb{R}^{d_{model} \times d_v}$: Value projection for head $i$
			\item $\mathbf{W}^O \in \mathbb{R}^{hd_v \times d_{model}}$: Output projection
		\end{itemize}
	\end{definition}
	
	Each head $i$ computes:
	\begin{equation}
		\text{head}_i = \text{softmax}\left(\frac{\mathbf{Q}\mathbf{W}_i^Q(\mathbf{K}\mathbf{W}_i^K)^T}{\sqrt{d_k}}\right)\mathbf{V}\mathbf{W}_i^V
	\end{equation}
	
	The outputs from all heads, each of dimension $n \times d_v$, are concatenated along the feature dimension to produce a matrix of dimension $n \times (hd_v)$, which is then linearly projected to dimension $n \times d_{model}$.
	
	\subsection{Dimension Choice}
	
	Typically, dimensions are chosen such that:
	\begin{equation}
		d_k = d_v = \frac{d_{model}}{h}
	\end{equation}
	
	With this choice, $hd_v = d_{model}$, and the total computational cost of multi-head attention is comparable to single-head attention with full dimensionality.
	
	\begin{example}
		For BERT base~\cite{devlin2019bert}: $d_{model} = 768$, $h = 12$, so $d_k = d_v = 64$.

		For GPT-3~\cite{brown2020language}: $d_{model} = 12288$, $h = 96$, so $d_k = d_v = 128$.
	\end{example}
	
	\section{Theoretical Analysis}
	
	\subsection{Representation Capacity}
	
	\begin{theorem}[Subspace Decomposition]
		Multi-head attention with $h$ heads can be viewed as learning $h$ different representation subspaces, where each head operates in a subspace of dimension $d_k = d_{model}/h$.
	\end{theorem}
	
	The projection matrices $\mathbf{W}_i^Q, \mathbf{W}_i^K, \mathbf{W}_i^V$ map the input into a lower-dimensional subspace where attention is computed. Different heads can learn to extract different features by focusing on different subspaces.
	
	\begin{theorem}[Expressiveness]
		Multi-head attention can represent a broader class of functions than single-head attention. Specifically, multi-head attention can simultaneously encode multiple relationships between positions, whereas single-head attention is limited to encoding a single weighted aggregation pattern.
	\end{theorem}
	
	\begin{trivlist}
		\item[\hskip\labelsep\bfseries Proof Sketch.]
		Consider a single-head attention mechanism. The output at position $i$ is:
		\begin{equation}
			\mathbf{y}_i = \sum_{j=1}^{n} \alpha_{ij} \mathbf{v}_j
		\end{equation}
		
		where $\alpha_{ij}$ is a scalar weight. The output is constrained to lie in the span of the value vectors $\{\mathbf{v}_j\}$.
		
		With multi-head attention, the output is:
		\begin{equation}
			\mathbf{y}_i = \sum_{k=1}^{h} \mathbf{W}^O_{k} \left(\sum_{j=1}^{n} \alpha_{ij}^{(k)} \mathbf{v}_j^{(k)}\right)
		\end{equation}
		
		where $\alpha_{ij}^{(k)}$ and $\mathbf{v}_j^{(k)}$ are the attention weights and values for head $k$. The output is now a weighted combination of outputs from different subspaces, allowing more flexible representations.
		
		Each head can learn different attention patterns $\{\alpha_{ij}^{(k)}\}_{i,j}$. For instance:
		\begin{itemize}
			\item Head 1 might learn to attend to adjacent positions (local context)
			\item Head 2 might learn to attend to syntactically related positions
			\item Head 3 might learn to attend to semantically similar positions
		\end{itemize}
		
		The final output projection $\mathbf{W}^O$ learns how to combine these different types of information.
	\end{trivlist}
	
	\subsection{Computational Complexity}
	
	\begin{proposition}[Complexity of Multi-Head Attention]
		For input sequence length $n$, model dimension $d_{model}$, and $h$ heads with $d_k = d_v = d_{model}/h$, multi-head attention has complexity:
		\begin{equation}
			O(n^2 d_{model} + n d_{model}^2)
		\end{equation}
		which is the same as single-head attention with full dimensionality.
	\end{proposition}
	
	\begin{trivlist}
		\item[\hskip\labelsep\bfseries Proof.]
		For each head $i$:
		\begin{enumerate}
			\item Computing $\mathbf{Q}\mathbf{W}_i^Q$, $\mathbf{K}\mathbf{W}_i^K$, $\mathbf{V}\mathbf{W}_i^V$: $3 \times O(n d_{\text{model}} d_k) = O(n d_{\text{model}}^2 / h)$
			\item Computing attention: $O(n^2 d_k) = O(n^2 d_{\text{model}} / h)$
		\end{enumerate}
		
		For all $h$ heads:
		\begin{equation}
			h \times \left(O\left(\frac{n d_{\text{model}}^2}{h}\right) + O\left(\frac{n^2 d_{\text{model}}}{h}\right)\right) = O(n d_{\text{model}}^2 + n^2 d_{\text{model}})
		\end{equation}
		
		The final output projection $\text{Concat}(\ldots)\mathbf{W}^O$ costs $O(n d_{\text{model}}^2)$.
		
		Total: $O(n^2 d_{\text{model}} + n d_{\text{model}}^2)$, same as single-head attention.
	\end{trivlist}
	
	This result shows that multi-head attention increases model capacity without increasing computational cost, provided dimensions are scaled appropriately.
	
	\section{Implementation and Optimization}
	
	\subsection{Efficient Implementation}
	
	Multi-head attention can be implemented efficiently using reshaped matrix operations that allow parallel computation of all heads.
	
	\begin{algorithm}[H]
		\caption{Efficient Multi-Head Attention}
		\begin{algorithmic}[1]
			\Require Inputs $\mathbf{Q}, \mathbf{K}, \mathbf{V} \in \mathbb{R}^{n \times d_{model}}$, heads $h$, dimensions $d_k = d_v = d_{model}/h$
			\Ensure Output $\mathbf{Y} \in \mathbb{R}^{n \times d_{model}}$
			\State Compute combined projections:
			\State \quad $\mathbf{Q}' \gets \mathbf{Q}\mathbf{W}^Q$ where $\mathbf{W}^Q \in \mathbb{R}^{d_{model} \times d_{model}}$
			\State \quad $\mathbf{K}' \gets \mathbf{K}\mathbf{W}^K$ where $\mathbf{W}^K \in \mathbb{R}^{d_{model} \times d_{model}}$
			\State \quad $\mathbf{V}' \gets \mathbf{V}\mathbf{W}^V$ where $\mathbf{W}^V \in \mathbb{R}^{d_{model} \times d_{model}}$
			\State Reshape for heads:
			\State \quad $\mathbf{Q}' \gets \text{reshape}(\mathbf{Q}', [n, h, d_k])$
			\State \quad $\mathbf{K}' \gets \text{reshape}(\mathbf{K}', [n, h, d_k])$
			\State \quad $\mathbf{V}' \gets \text{reshape}(\mathbf{V}', [n, h, d_v])$
			\State Transpose to $[h, n, d_k]$ for parallel head computation
			\State Compute attention for all heads in parallel:
			\State \quad $\mathbf{S} \gets \frac{\mathbf{Q}'\mathbf{K}'^T}{\sqrt{d_k}}$ \Comment{Shape: $[h, n, n]$}
			\State \quad $\mathbf{A} \gets \text{softmax}(\mathbf{S})$ \Comment{Row-wise for each head}
			\State \quad $\mathbf{O} \gets \mathbf{A}\mathbf{V}'$ \Comment{Shape: $[h, n, d_v]$}
			\State Transpose back to $[n, h, d_v]$ and reshape to $[n, hd_v]$
			\State Apply output projection:
			\State \quad $\mathbf{Y} \gets \mathbf{O}\mathbf{W}^O$
			\State \Return $\mathbf{Y}$
		\end{algorithmic}
	\end{algorithm}
	
	This implementation uses a single large matrix multiplication for all head projections (steps 1-3), rather than $h$ separate multiplications. The reshaping operations (steps 4-6) are typically just view operations that don't copy data.
	
	The parallel computation of attention for all heads (steps 7-9) exploits the batch dimension of tensor operations, allowing modern GPUs to compute all heads simultaneously.
	
	\subsection{Memory Optimization}
	
	For each head, we must store the attention weight matrix $\mathbf{A}_i \in \mathbb{R}^{n \times n}$ during training for backpropagation. With $h$ heads, total attention memory is:
	\begin{equation}
		M_{\text{attn}} = h \times n^2 \times \text{bytes\_per\_element}
	\end{equation}
	
	For long sequences, this becomes prohibitive. Several optimization strategies exist:
	
	\textbf{Gradient Checkpointing}: Don't store attention weights during forward pass. Recompute them during the backward pass from saved inputs. Trades computation for memory.
	
	\textbf{Memory-Efficient Attention}: Algorithms that compute attention gradients without materializing the full $n \times n$ matrices. These fuse softmax and matrix multiplication operations.
	
	\textbf{Sparse Attention}: Restrict each position to attend to only a subset of positions, reducing the attention matrix from $O(n^2)$ to $O(n \sqrt{n})$ or $O(n \log n)$ depending on the pattern.
	
	\section{Empirical Observations}
	
	\subsection{Head Specialization}
	
	Empirical analysis of trained Transformer models reveals that different heads learn distinct attention patterns:
	
	\begin{example}[Linguistic Phenomena in BERT]
		Analysis of BERT's attention heads~\cite{clark2019does} shows:
		\begin{itemize}
			\item \textbf{Positional heads}: Some heads attend primarily to the next or previous position, capturing sequential structure.
			
			\item \textbf{Syntactic heads}: Certain heads learn to follow dependency relationships, with attention patterns that resemble syntactic parse trees.
			
			\item \textbf{Semantic heads}: Some heads attend to semantically related words regardless of position, suggesting they capture semantic similarity.
			
			\item \textbf{Delimiter heads}: Heads that attend to special tokens like [SEP] and [CLS], using them as information hubs.
		\end{itemize}
	\end{example}
	
	\subsection{Layer-Wise Patterns}
	
	The role of multi-head attention changes across layers:
	
	\textbf{Early Layers}: Tend to focus on local, surface-level features. Attention patterns are often local (high weights near the diagonal) or positional.
	
	\textbf{Middle Layers}: Develop more complex patterns, including syntactic relationships and longer-range dependencies.
	
	\textbf{Late Layers}: Often exhibit more diffuse attention patterns, aggregating information from many positions. Some heads develop task-specific patterns related to the final prediction objective.
	
	\subsection{Head Importance}
	
	Not all heads contribute equally. Pruning experiments show:
	\begin{itemize}
		\item Many heads can be removed with minimal performance degradation
		\item Some heads are critical, and their removal significantly hurts performance
		\item Head importance varies by task and domain
	\end{itemize}
	
	This redundancy suggests that multi-head attention may be over-parameterized, motivating research into more efficient variants.
	
	\chapter{The Transformer Architecture}
	
	The Transformer architecture~\cite{vaswani2017attention} represents a complete neural network built entirely from attention mechanisms, eliminating recurrence and convolution. This chapter provides a comprehensive treatment of all components, with complete mathematical derivations.
	
	\section{Architecture Overview}
	
	The Transformer follows an encoder-decoder structure. Both encoder and decoder consist of stacked layers, each containing multi-head attention mechanisms and position-wise feed-forward networks, with residual connections and layer normalization.
	
	\subsection{Encoder Architecture}
	
	The encoder processes the input sequence and produces a sequence of contextual representations. It consists of $N$ identical layers (typically $N = 6$ or $N = 12$).
	
	Each encoder layer contains two sub-layers:
	\begin{enumerate}[leftmargin=*]
		\item \textbf{Multi-head self-attention mechanism}: Allows each position to attend to all positions in the encoder's previous layer.
		
		\item \textbf{Position-wise fully connected feed-forward network}: Applies the same feed-forward network independently to each position.
	\end{enumerate}
	
	Each sub-layer employs a residual connection followed by layer normalization:
	\begin{equation}
		\text{Output} = \text{LayerNorm}(\mathbf{x} + \text{Sublayer}(\mathbf{x}))
	\end{equation}
	
	where $\text{Sublayer}(\mathbf{x})$ is the function implemented by the sub-layer (either attention or feed-forward network).
	
	\subsection{Decoder Architecture}
	
	The decoder generates the output sequence one element at a time, conditioned on previously generated elements and the encoder outputs. It also consists of $N$ identical layers.
	
	Each decoder layer contains three sub-layers:
	\begin{enumerate}[leftmargin=*]
		\item \textbf{Masked multi-head self-attention}: Self-attention over the decoder's previous layer, with masking to prevent positions from attending to subsequent positions.
		
		\item \textbf{Multi-head encoder-decoder attention}: Queries come from the previous decoder layer, while keys and values come from the encoder output.
		
		\item \textbf{Position-wise feed-forward network}: Same as in the encoder.
	\end{enumerate}
	
\begin{figure}[h]
		\centering
		\begin{tikzpicture}[
			scale=0.8,
			every node/.style={font=\small},
			>=Stealth,
			thick,
			component/.style={draw, thick, fill=#1, minimum width=3.2cm, minimum height=0.8cm, align=center},
			container/.style={draw, thick, fill=anthropicCream!30, rounded corners=6pt},
			residual/.style={->, anthropicOrange, dashed, very thick, shorten >=2pt, shorten <=2pt}
			]
			
			\node[font=\bfseries\large, color=anthropicDeepOrange, anchor=west] at (-9, 10.7) {Encoder};
			
			\node[component=anthropicLightOrange!40] (enc_in) at (-8, 9.5) {Input Embedding};
			\node[component=anthropicBeige!60, minimum height=0.6cm] (enc_pos) at (-8, 8) {+ Positional Encoding};
			\draw[->, anthropicBrown, very thick] (enc_in) -- (enc_pos);
			
			\node[container, minimum width=4.5cm, minimum height=5cm] (enc_box) at (-8, 3.8) {};
			\node[font=\bfseries, color=anthropicOrange, anchor=south] at (-7.9, 6.8) {Encoder Layer $\times N$};
			
			\node[component=anthropicOrange!30] (enc_attn) at (-8, 6) {Multi-Head\\Self-Attention};
			\node[component=anthropicBeige, minimum height=0.6cm] (enc_norm1) at (-8, 4.5) {Add \& Norm};
			\node[component=anthropicTan!50] (enc_ff) at (-8, 3) {Feed Forward};
			\node[component=anthropicBeige, minimum height=0.6cm] (enc_norm2) at (-8, 1.5) {Add \& Norm};
			
			\draw[->, anthropicBrown, very thick] (enc_pos) -- (enc_attn);
			\draw[->, anthropicBrown, very thick] (enc_attn) -- (enc_norm1);
			\draw[->, anthropicBrown, very thick] (enc_norm1) -- (enc_ff);
			\draw[->, anthropicBrown, very thick] (enc_ff) -- (enc_norm2);
			
			\draw[residual] (enc_attn.west) -- ++(-0.5,0) |- (enc_norm1.west);
			\draw[residual] (enc_ff.west) -- ++(-0.5,0) |- (enc_norm2.west);
			
			\node[below=0.5cm of enc_norm2, font=\footnotesize, color=anthropicBrown] {Encoder Output};
			
			\node[font=\bfseries\large, color=anthropicDeepOrange, anchor=west] at (-1.1, 10.7) {Decoder};
			
			\node[component=anthropicLightOrange!40] (dec_in) at (0, 9.8) {Output Embedding};
			\node[component=anthropicBeige!60, minimum height=0.6cm] (dec_pos) at (0, 8.5) {+ Positional Encoding};
			\draw[->, anthropicBrown, very thick] (dec_in) -- (dec_pos);
			
			\node[container, minimum width=4.6cm, minimum height=9.8cm] (dec_box) at (0, 1.25) {};
			\node[font=\bfseries, color=anthropicOrange, anchor=south] at (0.1, 7.35) {Decoder Layer $\times N$};
			
			\node[component=anthropicDeepOrange!30] (dec_masked) at (0, 6.5) {Masked Multi-Head\\Self-Attention};
			\node[component=anthropicBeige, minimum height=0.6cm] (dec_norm1) at (0, 5) {Add \& Norm};
			\node[component=anthropicOrange!30] (dec_cross) at (0, 3.5) {Multi-Head\\Cross-Attention};
			\node[component=anthropicBeige, minimum height=0.6cm] (dec_norm2) at (0, 2) {Add \& Norm};
			\node[component=anthropicTan!50] (dec_ff) at (0, 0.5) {Feed Forward};
			\node[component=anthropicBeige, minimum height=0.6cm] (dec_norm3) at (0, -1) {Add \& Norm};
			
			\draw[->, anthropicBrown, very thick] (dec_pos) -- (dec_masked);
			\draw[->, anthropicBrown, very thick] (dec_masked) -- (dec_norm1);
			\draw[->, anthropicBrown, very thick] (dec_norm1) -- (dec_cross);
			\draw[->, anthropicBrown, very thick] (dec_cross) -- (dec_norm2);
			\draw[->, anthropicBrown, very thick] (dec_norm2) -- (dec_ff);
			\draw[->, anthropicBrown, very thick] (dec_ff) -- (dec_norm3);
			
			\draw[residual] (dec_masked.east) -- ++(0.5,0) |- (dec_norm1.east);
			\draw[residual] (dec_cross.east) -- ++(0.5,0) |- (dec_norm2.east);
			\draw[residual] (dec_ff.east) -- ++(0.5,0) |- (dec_norm3.east);
			
			\draw[residual] 
			(enc_norm2.east) -- ++(1,0) 
			-- ++(0,2) 
			-- (dec_cross.west);
			
			\node[font=\footnotesize, color=anthropicBrown, anchor=east] at (-3.5, 3.0) {K, V};
			
			\draw[residual, dotted] 
			(dec_norm1.west) -- ++(-1,0) 
			-- ++(0,-0.5) 
			-- (dec_cross.west);
			
			\node[font=\footnotesize, color=anthropicBrown, anchor=east] at (-3.0, 4.8) {Q};
			
			\node[component=anthropicLightOrange!60] (linear) at (0, -2.5) {Linear};
			\draw[->, anthropicBrown, very thick] (dec_norm3) -- (linear);
			
			\node[component=anthropicOrange!40] (softmax) at (0, -4) {Softmax};
			\draw[->, anthropicBrown, very thick] (linear) -- (softmax);
			
			\node[below=0.5cm of softmax, font=\footnotesize, color=anthropicBrown] {Output Probabilities};

		\end{tikzpicture}
		\caption{Complete Transformer encoder-decoder architecture. Orange dashed lines represent residual connections. The encoder processes the input sequence, and the decoder generates the output sequence conditioned on encoder outputs (as Keys and Values) and previous decoder outputs (as Query in cross-attention).}
		\label{fig:transformer_clean}
	\end{figure}
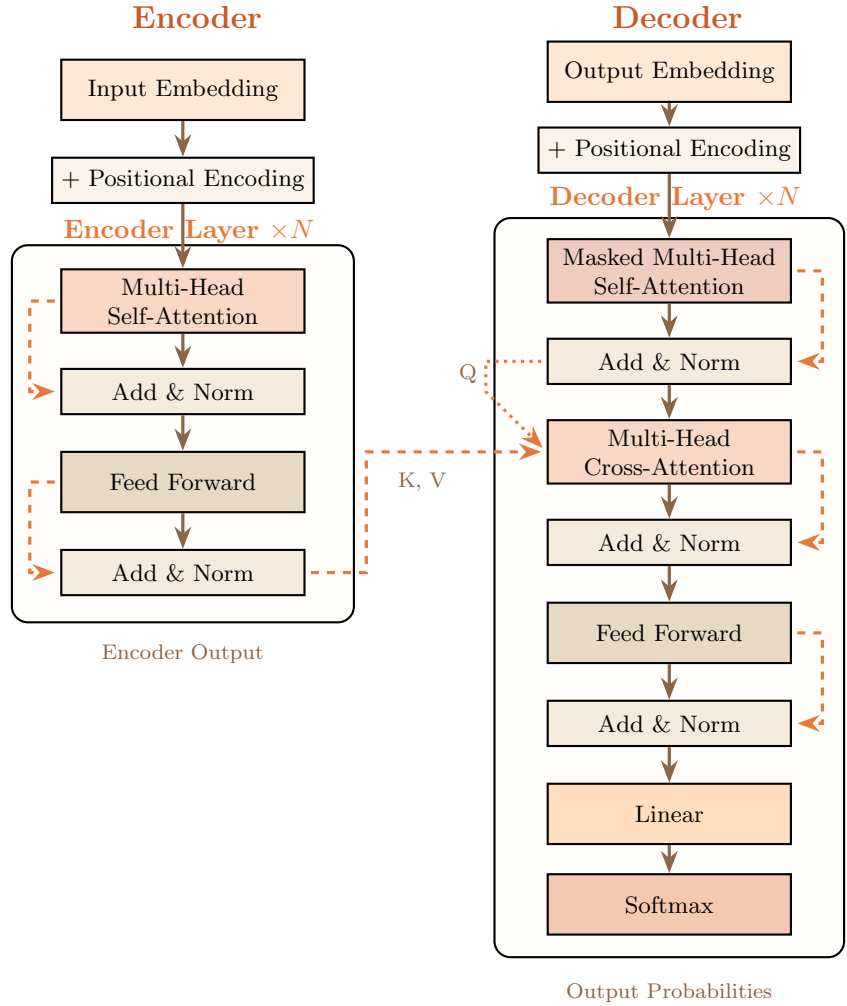

	Like the encoder, each sub-layer employs residual connections and layer normalization. The complete architecture is shown in \autoref{fig:transformer_clean}.

	\section{Positional Encoding}
	
	Since self-attention is permutation-equivariant, the Transformer has no inherent notion of sequence order. To incorporate positional information, the Transformer adds positional encodings to the input embeddings.
	
	\subsection{Sinusoidal Positional Encoding}
	
	The original Transformer uses sinusoidal functions of different frequencies:
	
	\begin{definition}[Sinusoidal Positional Encoding]
		For position $\text{pos} \in \{0, 1, \ldots, n-1\}$ and dimension $i \in \{0, 1, \ldots, d_{model}-1\}$:
		
		\begin{align}
			PE_{(\text{pos}, 2i)} &= \sin\left(\frac{\text{pos}}{10000^{2i/d_{model}}}\right) \label{eq:pe_sin}\\
			PE_{(\text{pos}, 2i+1)} &= \cos\left(\frac{\text{pos}}{10000^{2i/d_{model}}}\right) \label{eq:pe_cos}
		\end{align}
		
		where even dimensions use sine and odd dimensions use cosine.
	\end{definition}
	
	The wavelength of each dimension forms a geometric progression from $2\pi$ to $10000 \cdot 2\pi$.
	
	\subsection{Properties of Sinusoidal Encodings}
	
	\begin{proposition}[Relative Position Representation]
		For any fixed offset $k$, the positional encoding at position $\text{pos} + k$ can be represented as a linear function of the positional encoding at position $\text{pos}$.
	\end{proposition}
	
	\begin{trivlist}
		\item[\hskip\labelsep\bfseries Proof.]
		Using trigonometric identities:
		\begin{align}
			\sin(\alpha + \beta) &= \sin\alpha\cos\beta + \cos\alpha\sin\beta, \\
			\cos(\alpha + \beta) &= \cos\alpha\cos\beta - \sin\alpha\sin\beta.
		\end{align}
		
		Let $\omega_i = 1 / 10000^{2i/d_{\text{model}}}$. Then:
		\begin{align}
			PE_{(\text{pos}+k, 2i)} &= \sin(\omega_i(\text{pos}+k)) \\
			&= \sin(\omega_i \text{pos})\cos(\omega_i k) + \cos(\omega_i \text{pos})\sin(\omega_i k) \\
			&= PE_{(\text{pos}, 2i)}\cos(\omega_i k) + PE_{(\text{pos}, 2i+1)}\sin(\omega_i k).
		\end{align}
		
		Similarly, for the cosine term. This shows that $PE_{\text{pos}+k}$ can be expressed as a linear combination of $PE_{\text{pos}}$ with coefficients depending only on $k$ and $i$, not on $\text{pos}$.
	\end{trivlist}
	
	This property allows the model to attend to relative positions by learning appropriate linear transformations. Alternative approaches that directly encode relative position information into the attention mechanism have also been explored~\cite{shaw2018self}.
	
	\begin{insight}
		The sinusoidal encoding provides a unique encoding for each position while maintaining useful mathematical properties. The model can potentially learn to attend by relative position because $PE_{\text{pos}+k}$ is a linear function of $PE_{\text{pos}}$.
	\end{insight}
	
	\subsection{Learned Positional Embeddings}
	
	An alternative to sinusoidal encodings is to learn positional embeddings as parameters:
	
	\begin{equation}
		PE_{\text{pos}} = \mathbf{E}_{\text{pos}}
	\end{equation}
	
	where $\mathbf{E} \in \mathbb{R}^{n_{\max} \times d_{model}}$ is a learned embedding matrix and $n_{\max}$ is the maximum sequence length.
	
	\textbf{Advantages}:
	\begin{itemize}
		\item Can potentially adapt better to specific tasks
		\item No assumptions about the form of positional information
	\end{itemize}
	
	\textbf{Disadvantages}:
	\begin{itemize}
		\item Cannot extrapolate to sequences longer than seen during training
		\item Requires learning $O(n_{\max} d_{model})$ additional parameters
	\end{itemize}
	
	Empirical results show that learned and sinusoidal positional encodings achieve similar performance, though sinusoidal encodings have the theoretical advantage of being able to extrapolate to longer sequences.
	
	\section{Feed-Forward Networks}
	
	Each layer includes a position-wise fully connected feed-forward network that is applied identically and independently to each position.
	
	\begin{definition}[Position-Wise Feed-Forward Network]
		The feed-forward network consists of two linear transformations with a ReLU activation:
		
		\begin{equation}
			\text{FFN}(\mathbf{x}) = \max(0, \mathbf{x}\mathbf{W}_1 + \mathbf{b}_1)\mathbf{W}_2 + \mathbf{b}_2
		\end{equation}
		
		where:
		\begin{itemize}
			\item $\mathbf{W}_1 \in \mathbb{R}^{d_{model} \times d_{ff}}$: First layer weights
			\item $\mathbf{b}_1 \in \mathbb{R}^{d_{ff}}$: First layer bias
			\item $\mathbf{W}_2 \in \mathbb{R}^{d_{ff} \times d_{model}}$: Second layer weights
			\item $\mathbf{b}_2 \in \mathbb{R}^{d_{model}}$: Second layer bias
		\end{itemize}
		
		Typically, $d_{ff} = 4d_{model}$.
	\end{definition}
	
	The feed-forward network is "position-wise" because the same parameters are applied to each position independently. For an input sequence $\mathbf{X} \in \mathbb{R}^{n \times d_{model}}$, the output is:
	
	\begin{equation}
		[\text{FFN}(\mathbf{X})]_i = \text{FFN}(\mathbf{x}_i)
	\end{equation}
	
	where $\mathbf{x}_i$ is row $i$ of $\mathbf{X}$.
	
	\subsection{Role and Interpretation}
	
	The feed-forward network serves several purposes:
	
	\textbf{Increased Capacity}: The expansion to $d_{ff} = 4d_{model}$ dimensions provides additional representational capacity. The first layer projects to a higher dimension, and the second projects back.
	
	\textbf{Non-Linearity}: The ReLU activation introduces non-linearity, allowing the network to learn complex functions. Without this, stacking linear layers would be equivalent to a single linear layer.
	
	\textbf{Position-Specific Processing}: While attention aggregates information across positions, the feed-forward network processes each position independently, allowing position-specific transformations.
	
	\subsection{Alternative Activations}
	
	While ReLU is standard, other activations have been explored:
	
	\textbf{GELU (Gaussian Error Linear Unit)}:
	\begin{equation}
		\text{GELU}(x) = x \Phi(x)
	\end{equation}
	where $\Phi$ is the cumulative distribution function of the standard normal distribution. GELU is used in BERT and GPT models.
	
	\textbf{Swish}:
	\begin{equation}
		\text{Swish}(x) = x \cdot \sigma(\beta x)
	\end{equation}
	where $\sigma$ is the sigmoid function and $\beta$ is a learned or fixed parameter.
	
	These smooth activations can provide better gradient flow compared to ReLU.
	
	\section{Layer Normalization}

	Layer normalization~\cite{ba2016layer} normalizes activations across the feature dimension for each sample independently.
	
	\begin{definition}[Layer Normalization]
		For input $\mathbf{x} \in \mathbb{R}^{d}$, layer normalization computes:
		
		\begin{align}
			\mu &= \frac{1}{d}\sum_{i=1}^{d} x_i \label{eq:ln_mean}\\
			\sigma^2 &= \frac{1}{d}\sum_{i=1}^{d} (x_i - \mu)^2 \label{eq:ln_var}\\
			\hat{x}_i &= \frac{x_i - \mu}{\sqrt{\sigma^2 + \epsilon}} \label{eq:ln_normalize}\\
			\text{LayerNorm}(\mathbf{x})_i &= \gamma \hat{x}_i + \beta \label{eq:ln_scale}
		\end{align}
		
		where $\epsilon$ is a small constant (e.g., $10^{-6}$) for numerical stability, and $\gamma, \beta \in \mathbb{R}^{d}$ are learned parameters that allow the network to scale and shift the normalized values.
	\end{definition}
	
	\subsection{Properties and Benefits}
	
	\textbf{Normalization Across Features}: Unlike batch normalization, which normalizes across the batch dimension, layer normalization normalizes across features for each sample independently. This makes it suitable for variable-length sequences where batch statistics may be unreliable.
	
	\textbf{Training Stabilization}: Layer normalization reduces internal covariate shift, stabilizing the distribution of layer inputs and enabling faster training with higher learning rates.
	
	\textbf{Gradient Flow}: By normalizing activations, layer normalization helps maintain gradients within a reasonable range, facilitating the training of deep networks.
	
	\subsection{Placement: Pre-Norm vs Post-Norm}
	
	Two variants exist for placing layer normalization relative to residual connections:
	
	\textbf{Post-Norm} (Original Transformer):
	\begin{equation}
		\mathbf{y} = \text{LayerNorm}(\mathbf{x} + \text{Sublayer}(\mathbf{x}))
	\end{equation}
	
	\textbf{Pre-Norm}:
	\begin{equation}
		\mathbf{y} = \mathbf{x} + \text{Sublayer}(\text{LayerNorm}(\mathbf{x}))
	\end{equation}
	
	\begin{keypoint}
		Pre-norm has become more common in recent large models (e.g., GPT-3~\cite{brown2020language}, PaLM~\cite{chowdhery2022palm}) because it:
		\begin{itemize}
			\item Facilitates training of very deep networks
			\item Reduces gradient magnitudes, improving stability
			\item Often converges faster in practice
		\end{itemize}
		
		However, post-norm can achieve slightly better final performance with careful tuning.
	\end{keypoint}
	
	\section{Residual Connections}
	
	Residual connections~\cite{he2016deep}, introduced for computer vision, are crucial for training deep Transformers.
	
	\subsection{Formulation}
	
	Each sub-layer in the Transformer employs a residual connection:
	\begin{equation}
		\mathbf{y} = \mathbf{x} + \text{Sublayer}(\mathbf{x})
	\end{equation}
	
	where $\text{Sublayer}(\mathbf{x})$ is either attention or feed-forward computation.
	
	\subsection{Gradient Flow Analysis}
	
	\begin{theorem}[Residual Gradient Flow]
		Residual connections provide a direct path for gradients to flow from output to input, mitigating vanishing gradient problems in deep networks.
	\end{theorem}

	\begin{trivlist}
		\item[\hskip\labelsep\bfseries Proof.]
		Consider a layer with input $\mathbf{x}$ and output $\mathbf{y} = \mathbf{x} + F(\mathbf{x})$ where $F$ is the sub-layer function.
		
		During backpropagation, the gradient with respect to $\mathbf{x}$ is:
		\begin{equation}
			\frac{\partial \mathcal{L}}{\partial \mathbf{x}} = \frac{\partial \mathcal{L}}{\partial \mathbf{y}} \frac{\partial \mathbf{y}}{\partial \mathbf{x}} = \frac{\partial \mathcal{L}}{\partial \mathbf{y}} \left(I + \frac{\partial F(\mathbf{x})}{\partial \mathbf{x}}\right)
		\end{equation}
		
		where $I$ is the identity matrix. The identity term ensures that gradients can flow directly without being attenuated by the sub-layer, even if $\frac{\partial F(\mathbf{x})}{\partial \mathbf{x}}$ has small eigenvalues.
		
		For a network with $L$ layers, gradients flow as:
		\begin{equation}
			\frac{\partial \mathcal{L}}{\partial \mathbf{x}_0} = \frac{\partial \mathcal{L}}{\partial \mathbf{x}_L} \prod_{l=1}^{L} \left(I + \frac{\partial F_l(\mathbf{x}_{l-1})}{\partial \mathbf{x}_{l-1}}\right)
		\end{equation}
		
		The identity terms prevent the product from vanishing, as each factor is at least the identity.
	\end{trivlist}

	This gradient flow property enables training of very deep Transformers (e.g., 96 layers in large language models) without vanishing gradients.
	
	\chapter{Training and Optimization}
	
	Training Transformer models requires careful attention to optimization algorithms, learning rate schedules, regularization techniques, and loss functions. This chapter provides a comprehensive treatment of these topics with mathematical analysis.
	
	\section{Loss Functions}
	
	\subsection{Cross-Entropy Loss}
	
	For sequence-to-sequence tasks, the Transformer is trained to maximize the likelihood of the target sequence given the source sequence.
	
	\begin{definition}[Sequence Cross-Entropy Loss]
		For a sequence $\mathbf{y} = (y_1, \ldots, y_m)$ from vocabulary $\mathcal{V}$ of size $|V|$, the loss is:
		
		\begin{equation}
			\mathcal{L}(\theta) = -\sum_{t=1}^{m} \log P(y_t | y_{<t}, \mathbf{x}; \theta)
		\end{equation}
		
		where $\theta$ represents all model parameters, $\mathbf{x}$ is the source sequence, and $y_{<t} = (y_1, \ldots, y_{t-1})$ represents all previous target tokens.
	\end{definition}
	
	The model outputs logits $\mathbf{z}_t \in \mathbb{R}^{|V|}$ at each position $t$, which are converted to probabilities via softmax:
	
	\begin{equation}
		P(y_t = v | y_{<t}, \mathbf{x}; \theta) = \frac{\exp(z_{tv})}{\sum_{v'=1}^{|V|} \exp(z_{tv'})}
	\end{equation}
	
	The loss for position $t$ is then:
	\begin{equation}
		\mathcal{L}_t = -\log P(y_t | y_{<t}, \mathbf{x}; \theta) = -z_{t,y_t} + \log \sum_{v=1}^{|V|} \exp(z_{tv})
	\end{equation}
	
	\subsection{Label Smoothing}
	
	Label smoothing is a regularization technique that replaces hard targets with a mixture of the true label and a uniform distribution.
	
	\begin{definition}[Label Smoothing]
		Instead of using one-hot encoded targets, use smoothed targets:
		
		\begin{equation}
			\tilde{y}_{tv} = \begin{cases}
				1 - \epsilon + \frac{\epsilon}{|V|} & \text{if } v = y_t \\
				\frac{\epsilon}{|V|} & \text{otherwise}
			\end{cases}
		\end{equation}
		
		where $\epsilon$ is the smoothing parameter (typically $\epsilon = 0.1$).
	\end{definition}
	
	The loss becomes:
	\begin{equation}
		\mathcal{L}_t = -\sum_{v=1}^{|V|} \tilde{y}_{tv} \log P(y_t = v | y_{<t}, \mathbf{x}; \theta)
	\end{equation}
	
	\begin{proposition}[Effect of Label Smoothing]
		Label smoothing prevents the model from becoming overconfident in its predictions, encourages better calibration, and acts as a regularizer that can improve generalization.
	\end{proposition}
	
	The smoothed targets introduce entropy into the training signal, preventing the model from driving the logits of the correct class to $+\infty$ and all others to $-\infty$, which can lead to overfitting.
	
	\section{Optimization Algorithm}
	
	\subsection{Adam Optimizer}
	
	The original Transformer uses the Adam optimizer~\cite{kingma2014adam}, an adaptive learning rate method.
	
	\begin{algorithm}[H]
		\caption{Adam Optimization Algorithm}
		\begin{algorithmic}[1]
			\Require Learning rate $\alpha$, exponential decay rates $\beta_1, \beta_2$, small constant $\epsilon$
			\Require Initial parameters $\theta_0$
			\State Initialize first moment $\mathbf{m}_0 \gets 0$, second moment $\mathbf{v}_0 \gets 0$
			\State Initialize time step $t \gets 0$
			\While{not converged}
			\State $t \gets t + 1$
			\State Compute gradient: $\mathbf{g}_t \gets \nabla_\theta \mathcal{L}(\theta_{t-1})$
			\State Update first moment: $\mathbf{m}_t \gets \beta_1 \mathbf{m}_{t-1} + (1 - \beta_1)\mathbf{g}_t$
			\State Update second moment: $\mathbf{v}_t \gets \beta_2 \mathbf{v}_{t-1} + (1 - \beta_2)\mathbf{g}_t^2$
			\State Compute bias-corrected moments:
			\State \quad $\hat{\mathbf{m}}_t \gets \frac{\mathbf{m}_t}{1 - \beta_1^t}$
			\State \quad $\hat{\mathbf{v}}_t \gets \frac{\mathbf{v}_t}{1 - \beta_2^t}$
			\State Update parameters: $\theta_t \gets \theta_{t-1} - \alpha \frac{\hat{\mathbf{m}}_t}{\sqrt{\hat{\mathbf{v}}_t} + \epsilon}$
			\EndWhile
		\end{algorithmic}
	\end{algorithm}
	
	Standard hyperparameters: $\beta_1 = 0.9$, $\beta_2 = 0.999$, $\epsilon = 10^{-8}$.
	
	Adam adapts the learning rate for each parameter based on estimates of first and second moments of the gradients, allowing different learning rates for different parameters.
	
	\subsection{Learning Rate Schedule}
	
	The original Transformer employs a learning rate schedule that increases linearly during a warmup phase and then decreases.
	
	\begin{definition}[Warmup Learning Rate Schedule]
		The learning rate at step $t$ is:
		
		\begin{equation}
			\text{lr}(t) = d_{model}^{-0.5} \cdot \min(t^{-0.5}, t \cdot \text{warmup\_steps}^{-1.5})
		\end{equation}
		
		where warmup\_steps is the number of warmup steps (typically 4000-8000).
	\end{definition}
	
	This schedule has two phases:
	\begin{enumerate}
		\item \textbf{Warmup} ($t < \text{warmup\_steps}$): Learning rate increases linearly:
		\begin{equation}
			\text{lr}(t) = d_{model}^{-0.5} \cdot t \cdot \text{warmup\_steps}^{-1.5}
		\end{equation}
		
		\item \textbf{Decay} ($t \geq \text{warmup\_steps}$): Learning rate decreases as $t^{-0.5}$:
		\begin{equation}
			\text{lr}(t) = d_{model}^{-0.5} \cdot t^{-0.5}
		\end{equation}
	\end{enumerate}
	
	\begin{insight}
		The warmup phase is crucial for stable training. Without a warmup, Adam with a high initial learning rate can destabilize training, particularly in the first few iterations when second-moment estimates are inaccurate.
		
		The $d_{model}^{-0.5}$ scaling ensures that the effective learning rate is adjusted based on model dimensionality, preventing excessively large updates in high-dimensional models.
	\end{insight}
	
	\section{Regularization Techniques}
	
	\subsection{Dropout}
	
	Dropout~\cite{srivastava2014dropout} is applied at multiple points in the Transformer:
	
	\begin{definition}[Dropout]
		During training, dropout randomly sets elements of the input to zero with probability $p$, and scales remaining elements by $\frac{1}{1-p}$:
		
		\begin{equation}
			\text{Dropout}(\mathbf{x})_i = \begin{cases}
				0 & \text{with probability } p \\
				\frac{x_i}{1-p} & \text{with probability } 1-p
			\end{cases}
		\end{equation}
		
		During evaluation, dropout is turned off: $\text{Dropout}(\mathbf{x}) = \mathbf{x}$.
	\end{definition}
	
	Dropout is applied to:
	\begin{enumerate}
		\item Output of each sub-layer before the residual connection
		\item Attention weights after softmax
		\item Input embeddings
	\end{enumerate}
	
	Typical dropout rates: $p = 0.1$ for large datasets, $p = 0.3$ for smaller datasets.
	
	\subsection{Attention Dropout}
	
	Attention dropout is applied to the attention weight matrix after softmax:
	
	\begin{equation}
		\mathbf{y} = \text{Dropout}(\text{softmax}(\mathbf{S}))\mathbf{V}
	\end{equation}
	
	where $\mathbf{S}$ is the scaled score matrix. This prevents the model from relying too heavily on specific attention patterns.
	
	\subsection{Weight Decay}
	
	Weight decay (L2 regularization) adds a penalty term to the loss:
	
	\begin{equation}
		\mathcal{L}_{\text{total}} = \mathcal{L}_{\text{task}} + \frac{\lambda}{2}\sum_{i} \theta_i^2
	\end{equation}
	
	where $\lambda$ is the weight decay coefficient (typically $\lambda = 0.01$ to $0.0001$).
	
	In Adam, weight decay is often decoupled from the adaptive learning rate (AdamW):
	
	\begin{equation}
		\theta_t \gets \theta_{t-1} - \alpha \frac{\hat{\mathbf{m}}_t}{\sqrt{\hat{\mathbf{v}}_t} + \epsilon} - \alpha \lambda \theta_{t-1}
	\end{equation}
	
	\section{Gradient Computation and Backpropagation}
	
	\subsection{Backpropagation Through Attention}
	
	We derive the gradients for the attention mechanism to understand how errors propagate.
	
	For attention output:
	\begin{equation}
		\mathbf{y}_i = \sum_{j=1}^{n} \alpha_{ij} \mathbf{v}_j
	\end{equation}
	
	Given gradient $\frac{\partial \mathcal{L}}{\partial \mathbf{y}_i}$ from the next layer:
	
	\textbf{Gradient with respect to values}:
	\begin{equation}
		\frac{\partial \mathcal{L}}{\partial \mathbf{v}_j} = \sum_{i=1}^{m} \alpha_{ij} \frac{\partial \mathcal{L}}{\partial \mathbf{y}_i}
	\end{equation}
	
	\textbf{Gradient with respect to attention weights}:
	\begin{equation}
		\frac{\partial \mathcal{L}}{\partial \alpha_{ij}} = \frac{\partial \mathcal{L}}{\partial \mathbf{y}_i}^T \mathbf{v}_j
	\end{equation}
	
	\textbf{Gradient through softmax}: For the softmax function $\alpha_{ij} = \frac{\exp(s_{ij})}{\sum_k \exp(s_{ik})}$:
	
	\begin{equation}
		\frac{\partial \mathcal{L}}{\partial s_{ij}} = \sum_{k=1}^{n} \frac{\partial \mathcal{L}}{\partial \alpha_{ik}} \frac{\partial \alpha_{ik}}{\partial s_{ij}}
	\end{equation}
	
	The Jacobian of softmax is:
	\begin{equation}
		\frac{\partial \alpha_{ik}}{\partial s_{ij}} = \begin{cases}
			\alpha_{ij}(1 - \alpha_{ij}) & \text{if } k = j \\
			-\alpha_{ij}\alpha_{ik} & \text{if } k \neq j
		\end{cases}
	\end{equation}
	
	Therefore:
	\begin{equation}
		\frac{\partial \mathcal{L}}{\partial s_{ij}} = \alpha_{ij}\left(\frac{\partial \mathcal{L}}{\partial \alpha_{ij}} - \sum_{k=1}^{n} \alpha_{ik}\frac{\partial \mathcal{L}}{\partial \alpha_{ik}}\right)
	\end{equation}
	
	These gradients are then backpropagated through the query and key computations.
	
	\subsection{Gradient Clipping}
	
	To prevent exploding gradients, gradient clipping is often applied:
	
	\begin{equation}
		\mathbf{g} \gets \begin{cases}
			\mathbf{g} & \text{if } \|\mathbf{g}\| \leq \theta \\
			\frac{\theta}{\|\mathbf{g}\|}\mathbf{g} & \text{if } \|\mathbf{g}\| > \theta
		\end{cases}
	\end{equation}
	
	where $\theta$ is the clipping threshold (typically $\theta = 1.0$ or $5.0$).
	
	\section{Mixed Precision Training}
	
	Modern Transformers are often trained using mixed precision, combining 16-bit and 32-bit floating point computations.
	
	\subsection{FP16/FP32 Mixed Precision}
	
	\begin{keypoint}
		Mixed precision training:
		\begin{itemize}
			\item Stores model parameters in FP32
			\item Computes forward and backward passes in FP16
			\item Uses loss scaling to prevent underflow
			\item Updates parameters in FP32
		\end{itemize}
		
		Benefits:
		\begin{itemize}
			\item 2× memory savings for activations
			\item 2-3× speedup on GPUs with Tensor Cores
			\item Minimal impact on model quality with proper implementation
		\end{itemize}
	\end{keypoint}
	
	\textbf{Loss Scaling}: Multiply loss by a large constant (e.g., 1024 or 2048) before backpropagation to keep gradients in FP16 range:
	
	\begin{equation}
		\mathcal{L}_{\text{scaled}} = \mathcal{L} \cdot \text{scale}
	\end{equation}
	
	After computing gradients, divide by the scale factor before parameter updates:
	
	\begin{equation}
		\mathbf{g}_{\text{true}} = \frac{\mathbf{g}_{\text{scaled}}}{\text{scale}}
	\end{equation}
	
	\section{Distributed Training}
	
	Large Transformers require distributed training across multiple GPUs or machines.
	
	\subsection{Data Parallelism}
	
	Each GPU maintains a complete copy of the model and processes different batches. Gradients are averaged across GPUs before parameter updates.
	
	\textbf{Algorithm}:
	\begin{enumerate}
		\item Partition batch across GPUs
		\item Each GPU computes the forward pass on its partition
		\item Each GPU computes gradients via backpropagation
		\item All-reduce operation averages gradients across GPUs
		\item Each GPU updates its model parameters
	\end{enumerate}
	
	\subsection{Model Parallelism}
	
	For models too large to fit on a single GPU, different layers or attention heads are placed on different GPUs.
	
	\textbf{Tensor Parallelism}: Splits individual tensors (e.g., weight matrices) across GPUs, requiring communication during forward and backward passes.
	
	\textbf{Pipeline Parallelism}: Assigns different layers to different GPUs, processing micro-batches in a pipelined fashion to reduce idle time.
	
	\chapter{Conclusion and Future Directions}

	This comprehensive treatment has established the mathematical foundations of attention mechanisms and their implementation in the Transformer architecture. From the fundamental attention operation through multi-head attention, positional encoding, and training dynamics, we have developed a rigorous understanding of these powerful sequence modeling tools. Attention mechanisms represent a fundamental paradigm shift in how neural networks process sequential information, moving from fixed compression through recurrent states to flexible, learned aggregation patterns. The success of Transformers across natural language processing, computer vision, and multimodal domains demonstrates the broad applicability of this approach.

	\section{Achievements and Impact}

	The Transformer architecture has fundamentally reshaped modern machine learning. Large language models built on Transformer foundations, including GPT-3~\cite{brown2020language}, PaLM~\cite{chowdhery2022palm}, and LLaMA~\cite{touvron2023llama}, have demonstrated remarkable capabilities across diverse tasks through in-context learning and few-shot adaptation. These models have achieved performance levels that approach or exceed human capabilities on many benchmarks, while also revealing emergent properties not present in smaller-scale systems. Beyond language processing, attention mechanisms have enabled breakthroughs in protein structure prediction through AlphaFold~\cite{jumper2021highly}, revolutionized computer vision with Vision Transformers~\cite{dosovitskiy2020image}, and facilitated multimodal understanding through models like CLIP~\cite{radford2021learning} and Flamingo~\cite{alayrac2022flamingo}. The versatility of attention as a general-purpose mechanism for relationship modeling has proven applicable across scientific domains, from molecular dynamics to climate modeling.

	\section{Efficiency and Scalability Challenges}

	Despite these successes, the quadratic complexity of standard attention with respect to sequence length remains a fundamental limitation. For sequences of length $n$, computing all pairwise attention scores requires $O(n^2)$ operations and memory, making it computationally prohibitive for very long contexts. This has motivated extensive research into efficient attention variants. Sparse attention patterns, as explored in Sparse Transformers~\cite{child2019generating} and Longformer~\cite{beltagy2020longformer}, reduce complexity by restricting attention to local windows and selected global positions. These methods achieve linear or near-linear complexity while maintaining performance on tasks requiring long-range dependencies. Big Bird~\cite{zaheer2020big} combines local, global, and random attention patterns, proving theoretically that sparse attention can approximate full attention while dramatically reducing computational cost.

	Low-rank approximation methods offer another approach to efficiency. Linformer~\cite{wang2020linformer} projects keys and values to lower dimensions, reducing complexity to $O(nk)$ where $k \ll n$. Linear attention methods~\cite{katharopoulos2020transformers} and the Performer~\cite{choromanski2020rethinking} use random feature approximations to estimate attention scores, achieving linear complexity through kernel methods. These approaches trade some expressiveness for computational efficiency, with empirical results suggesting the trade-off is often favorable. Recent work on Flash Attention~\cite{dao2022flashattention} demonstrates that algorithmic optimizations exploiting hardware characteristics can yield substantial speedups without sacrificing accuracy. By carefully managing memory access patterns and leveraging GPU memory hierarchy, Flash Attention reduces memory reads/writes while computing exact attention, achieving 2-4x speedups over standard implementations. This highlights that implementation details matter as much as algorithmic complexity for practical performance.

	\section{Theoretical Understanding}

	While the empirical success of Transformers is undeniable, the theoretical understanding of why and how they work remains incomplete. Recent theoretical work has begun addressing fundamental questions about Transformers' representational capacity and learning dynamics. Yun et al.~\cite{yun2019transformers} proved that Transformers with sufficient depth can represent any continuous sequence-to-sequence function, establishing universal approximation properties. However, these results require impractical depth, leaving open questions about what functions Transformers can efficiently learn in practice. Analysis of attention as performing similarity-based retrieval~\cite{elhage2021mathematical} provides intuition for how Transformers process information, but a rigorous characterization of learned representations remains challenging. Understanding training dynamics presents additional challenges. The interplay between attention layers, feed-forward networks, and residual connections creates complex optimization landscapes. Recent work on neural tangent kernels for Transformers~\cite{hron2020infinite} provides some theoretical grounding, but the gap between infinite-width theory and finite practical models limits applicability. Developing a theory that explains phenomena like in-context learning, where models adapt to new tasks from examples in their input without gradient updates, represents an important frontier.

	\section{Interpretability and Analysis}

	The attention weights in Transformers provide some interpretability, allowing inspection of which input positions influence each output. However, attention patterns often prove difficult to interpret meaningfully, particularly in deeper layers where representations become increasingly abstract. Research into Transformer interpretability has revealed both insights and limitations. Studies of attention patterns in trained models~\cite{clark2019does} have identified specialized attention heads that attend to specific syntactic relationships in language tasks, suggesting some degree of interpretable structure. However, other work~\cite{jain2019attention} demonstrates that attention weights do not always correspond to model behavior or feature importance, cautioning against over-interpreting attention as explanation. The distinction between attention as a mechanism and attention as explanation remains an active research question. Advances in mechanistic interpretability~\cite{elhage2021mathematical} aim to reverse-engineer learned algorithms in Transformers by analyzing internal representations and information flow. These approaches have successfully identified interpretable circuits for specific tasks, but scaling to large models trained on diverse data remains challenging. Developing robust methods for understanding, debugging, and controlling Transformer behavior represents a critical research direction for deploying these systems safely and reliably.

	\section{Future Directions}

	\subsection{Long-Context Modeling}

	Extending Transformers to handle extremely long contexts efficiently remains an important goal. Applications including book-length document understanding, long-form dialogue, and scientific literature analysis require processing sequences far exceeding current practical limits. Approaches combining hierarchical processing, memory mechanisms, and retrieval augmentation show promise for scaling to arbitrarily long contexts while maintaining computational tractability.

	\subsection{Multimodal Integration}

	While attention has proven effective for multimodal learning, better methods for integrating diverse data types are needed. Cross-modal attention mechanisms that capture fine-grained alignments between vision and language~\cite{alayrac2022flamingo} represent progress, but open questions remain about optimal architectures for three or more modalities, temporal alignment across modalities, and handling modality-specific structure.

	\subsection{Efficient Architectures}

	Hybrid architectures combining Transformers with other mechanisms may offer better efficiency-performance trade-offs. State space models like S4~\cite{gu2021efficiently} achieve competitive performance with linear complexity by replacing attention with structured linear operators. Understanding when attention is necessary versus when alternatives suffice could enable more efficient architectures tailored to specific tasks. Mixture-of-Experts approaches~\cite{fedus2022switch} offer another path to scaling by activating only subsets of model parameters for each input. Combining sparse attention with sparse expert routing may enable models with trillions of parameters while maintaining manageable computational costs.

	\subsection{Continual and Few-Shot Learning}

	While Transformers exhibit impressive in-context learning, improving their ability to efficiently acquire and retain new knowledge remains important. Better understanding and enhancing few-shot learning capabilities could reduce data requirements for adapting models to specialized domains. Developing architectures that support continual learning without catastrophic forgetting while maintaining computational efficiency represents a key challenge.

	\subsection{Environmental Considerations}

	The computational cost of training large Transformers raises important questions about energy consumption and environmental impact. Research into more efficient training methods, better transfer learning to reduce training from scratch, and specialized hardware for Transformer inference addresses these concerns. Developing metrics and best practices for sustainable AI development will become increasingly important as models scale.

	\section{Concluding Remarks}

	Attention mechanisms have transformed artificial intelligence by providing a flexible, powerful framework for modeling relationships in sequential and structured data. The mathematical foundations developed in this monograph establish the theoretical basis for understanding these mechanisms, while highlighting open questions and challenges. As research addresses efficiency, theoretical understanding, interpretability, and domain-specific applications, attention-based architectures will continue evolving. Whether through improved Transformer variants, hybrid approaches combining multiple mechanisms, or entirely new architectures inspired by attention principles, the core insight of learned, adaptive aggregation will remain central to progress in machine learning. The next decade promises continued innovation, building on the attention paradigm established over the past several years.

	\appendix

	\chapter{Notation and Mathematical Conventions}
	\label{app:notation}

	This appendix provides a comprehensive reference for the mathematical notation, symbols, and conventions used throughout this monograph.

	\section{General Notation}

	\subsection{Scalars, Vectors, and Matrices}

	\begin{table}[h]
		\centering
		\begin{tabular}{ll}
			\toprule
			\textbf{Symbol} & \textbf{Description} \\
			\midrule
			$a, b, c, \alpha, \beta$ & Scalars (lowercase italic) \\
			$\mathbf{x}, \mathbf{y}, \mathbf{z}$ & Vectors (lowercase bold) \\
			$\mathbf{A}, \mathbf{B}, \mathbf{W}$ & Matrices (uppercase bold) \\
			$x_i$ & $i$-th element of vector $\mathbf{x}$ \\
			$A_{ij}$ & Element at row $i$, column $j$ of matrix $\mathbf{A}$ \\
			$\mathbf{A}^T$ & Transpose of matrix $\mathbf{A}$ \\
			$\mathbf{A}^{-1}$ & Inverse of matrix $\mathbf{A}$ \\
			$\|\mathbf{x}\|$ & Euclidean (L2) norm of vector $\mathbf{x}$ \\
			$\|\mathbf{x}\|_p$ & $L_p$ norm of vector $\mathbf{x}$ \\
			$\langle \mathbf{x}, \mathbf{y} \rangle$ & Inner product of vectors $\mathbf{x}$ and $\mathbf{y}$ \\
			\bottomrule
		\end{tabular}
		\caption{Scalar, vector, and matrix notation}
		\label{tab:notation_basic}
	\end{table}

	\subsection{Sets and Spaces}

	\begin{table}[h]
		\centering
		\begin{tabular}{ll}
			\toprule
			\textbf{Symbol} & \textbf{Description} \\
			\midrule
			$\mathbb{R}$ & Set of real numbers \\
			$\mathbb{R}^d$ & $d$-dimensional Euclidean space \\
			$\mathbb{R}^{m \times n}$ & Space of $m \times n$ real matrices \\
			$\mathbb{N}$ & Set of natural numbers \\
			$\mathbb{Z}$ & Set of integers \\
			$[n]$ & Set $\{1, 2, \ldots, n\}$ \\
			$|S|$ & Cardinality of set $S$ \\
			\bottomrule
		\end{tabular}
		\caption{Sets and spaces}
		\label{tab:notation_sets}
	\end{table}

	\section{Attention-Specific Notation}

	\subsection{Sequences and Positions}

	\begin{table}[h]
		\centering
		\begin{tabular}{ll}
			\toprule
			\textbf{Symbol} & \textbf{Description} \\
			\midrule
			$n$ & Sequence length \\
			$d, d_{model}$ & Model/embedding dimension \\
			$d_k$ & Key/query dimension \\
			$d_v$ & Value dimension \\
			$\mathbf{x}_i$ & Input vector at position $i$ \\
			$\mathbf{X} \in \mathbb{R}^{n \times d}$ & Input sequence matrix \\
			$\mathbf{h}_i$ & Hidden state at position $i$ \\
			$\mathbf{H} \in \mathbb{R}^{n \times d}$ & Hidden state sequence matrix \\
			\bottomrule
		\end{tabular}
		\caption{Sequence notation}
		\label{tab:notation_sequences}
	\end{table}

	\subsection{Attention Components}

	\begin{table}[h]
		\centering
		\begin{tabular}{ll}
			\toprule
			\textbf{Symbol} & \textbf{Description} \\
			\midrule
			$\mathbf{q}_i$ & Query vector at position $i$ \\
			$\mathbf{k}_j$ & Key vector at position $j$ \\
			$\mathbf{v}_j$ & Value vector at position $j$ \\
			$\mathbf{Q} \in \mathbb{R}^{n \times d_k}$ & Query matrix \\
			$\mathbf{K} \in \mathbb{R}^{n \times d_k}$ & Key matrix \\
			$\mathbf{V} \in \mathbb{R}^{n \times d_v}$ & Value matrix \\
			$e_{ij}$ & Unnormalized attention score between $i$ and $j$ \\
			$\alpha_{ij}$ & Attention weight from position $i$ to $j$ \\
			$\mathbf{A} \in \mathbb{R}^{n \times n}$ & Attention weight matrix \\
			$\mathbf{c}_i$ & Context vector at position $i$ \\
			\bottomrule
		\end{tabular}
		\caption{Attention mechanism notation}
		\label{tab:notation_attention}
	\end{table}

	\subsection{Transformer Components}

	\begin{table}[h]
		\centering
		\begin{tabular}{ll}
			\toprule
			\textbf{Symbol} & \textbf{Description} \\
			\midrule
			$h$ & Number of attention heads \\
			$d_{ff}$ & Feed-forward network hidden dimension \\
			$L$ & Number of layers \\
			$\mathbf{W}^Q, \mathbf{W}^K, \mathbf{W}^V$ & Query, key, value projection matrices \\
			$\mathbf{W}^O$ & Output projection matrix \\
			$PE_{\text{pos}}$ & Positional encoding at position $\text{pos}$ \\
			$\text{LayerNorm}(\cdot)$ & Layer normalization function \\
			$\text{FFN}(\cdot)$ & Feed-forward network \\
			$\text{softmax}(\cdot)$ & Softmax activation function \\
			\bottomrule
		\end{tabular}
		\caption{Transformer architecture notation}
		\label{tab:notation_transformer}
	\end{table}

	\section{Functions and Operations}

	\subsection{Activation Functions}

	\begin{table}[h]
		\centering
		\begin{tabular}{ll}
			\toprule
			\textbf{Function} & \textbf{Definition} \\
			\midrule
			$\text{softmax}(\mathbf{x})_i$ & $\frac{\exp(x_i)}{\sum_{j} \exp(x_j)}$ \\
			$\text{ReLU}(x)$ & $\max(0, x)$ \\
			$\text{GELU}(x)$ & $x \Phi(x)$ where $\Phi$ is Gaussian CDF \\
			$\sigma(x)$ & $\frac{1}{1 + \exp(-x)}$ (sigmoid) \\
			$\tanh(x)$ & $\frac{\exp(x) - \exp(-x)}{\exp(x) + \exp(-x)}$ \\
			\bottomrule
		\end{tabular}
		\caption{Common activation functions}
		\label{tab:notation_activations}
	\end{table}

	\subsection{Mathematical Operations}

	\begin{table}[h]
		\centering
		\begin{tabular}{ll}
			\toprule
			\textbf{Symbol} & \textbf{Description} \\
			\midrule
			$\nabla_{\mathbf{x}} f$ & Gradient of $f$ with respect to $\mathbf{x}$ \\
			$\frac{\partial f}{\partial x}$ & Partial derivative of $f$ with respect to $x$ \\
			$\mathbb{E}[\cdot]$ & Expected value \\
			$\text{Var}[\cdot]$ & Variance \\
			$\odot$ & Element-wise (Hadamard) product \\
			$\oplus$ & Concatenation operation \\
			$\arg\max$ & Argument of the maximum \\
			\bottomrule
		\end{tabular}
		\caption{Mathematical operations}
		\label{tab:notation_operations}
	\end{table}

	\section{Complexity Notation}

	\begin{table}[h]
		\centering
		\begin{tabular}{ll}
			\toprule
			\textbf{Notation} & \textbf{Description} \\
			\midrule
			$O(f(n))$ & Big-O notation: asymptotic upper bound \\
			$\Omega(f(n))$ & Big-Omega: asymptotic lower bound \\
			$\Theta(f(n))$ & Big-Theta: asymptotic tight bound \\
			$o(f(n))$ & Little-o: strictly smaller asymptotic growth \\
			\bottomrule
		\end{tabular}
		\caption{Complexity notation}
		\label{tab:notation_complexity}
	\end{table}

	\section{Common Abbreviations}

	\begin{table}[h]
		\centering
		\begin{tabular}{ll}
			\toprule
			\textbf{Abbreviation} & \textbf{Full Term} \\
			\midrule
			MHA & Multi-Head Attention \\
			FFN & Feed-Forward Network \\
			LN & Layer Normalization \\
			PE & Positional Encoding \\
			LSTM & Long Short-Term Memory \\
			RNN & Recurrent Neural Network \\
			CNN & Convolutional Neural Network \\
			NLP & Natural Language Processing \\
			BERT & Bidirectional Encoder Representations from Transformers \\
			GPT & Generative Pre-trained Transformer \\
			\bottomrule
		\end{tabular}
		\caption{Common abbreviations}
		\label{tab:abbreviations}
	\end{table}

	\section{Conventions}

	\subsection{Indexing}
	\begin{itemize}
		\item Sequence positions are indexed from 1 to $n$, i.e., $i, j \in [n] = \{1, 2, \ldots, n\}$
		\item Vector and matrix indices start from 1 unless otherwise specified
		\item In code examples, 0-based indexing may be used following programming conventions
	\end{itemize}

	\subsection{Dimensions}
	\begin{itemize}
		\item Input sequences are represented as matrices $\mathbf{X} \in \mathbb{R}^{n \times d}$ where rows correspond to positions
		\item Matrix multiplications follow standard conventions: $\mathbf{AB}$ requires compatible dimensions
		\item Broadcasting rules follow NumPy/PyTorch conventions when discussing implementations
	\end{itemize}

	\subsection{Summation and Products}
	\begin{itemize}
		\item Summation notation: $\sum_{i=1}^{n}$ or $\sum_{i \in [n]}$ for sums over sequence positions
		\item When the range is clear from context, we may write $\sum_i$ or $\sum_j$
		\item Matrix products are denoted by juxtaposition: $\mathbf{AB}$
		\item Element-wise products use $\odot$: $\mathbf{A} \odot \mathbf{B}$
	\end{itemize}

	\subsection{Probability and Expectations}
	\begin{itemize}
		\item Random variables are denoted with uppercase letters: $X, Y$
		\item Probability distributions: $p(x)$ or $P(X = x)$
		\item Conditional probability: $p(y|x)$ or $P(Y = y | X = x)$
		\item Expected values: $\mathbb{E}_{x \sim p}[f(x)]$ or $\mathbb{E}[f(X)]$ when distribution is clear
	\end{itemize}

	For detailed references to specific notation in context, please see the relevant chapters. \autoref{tab:notation_attention} provides the core attention mechanism notation used extensively in Chapters 2--4, while \autoref{tab:notation_transformer} covers the Transformer-specific notation in Chapters 5--6.

	\backmatter

	\small
	\bibliographystyle{plain}
	\begin{multicols}{2}[\chapter*{References}\addcontentsline{toc}{chapter}{References}]
	\begingroup
	\renewcommand{\chapter}[2]{}
	\bibliography{references}
	\endgroup
	\end{multicols}

\end{document}